\documentclass[]{aero}
\newcommand{\mb}[1]{\mathbf{#1}}
\usepackage{siunitx}
\usepackage{amsmath}
\usepackage{amssymb}
\usepackage{tikz}
\usepackage{booktabs}  
\usepackage{array}     
\usepackage{makecell}  
\usepackage{caption}
\usepackage{subcaption}
\usepackage{svg}
\usepackage[capitalise, noabbrev]{cleveref}
\usepackage{siunitx}
\usepackage{float}
\usepackage{mathrsfs} 

\usepackage{bm} 
\renewcommand{\mb}[1]{\bm{#1}}

\usetikzlibrary{calc,patterns,angles,quotes}

\begin{document}
\title{Vision-Guided Optic Flow Navigation for Small Lunar Missions}
\author{Sean Cowan$^{1}$, Pietro Fanti$^{1}$, Leon B. S. Williams$^{1}$, Chit Hong Yam$^{2}$, Kaneyasu Asakuma$^{2}$, Yuichiro Nada$^{2}$, Dario Izzo$^{1}$\cor{}}
\maketitle  
\noindent\cor{} dario.izzo@esa.int \\
\\
\noindent\textit{1. Advanced Concepts Team, European Space Research and Technology Centre (ESTEC), Noordwijk, The Netherlands.\\}
\noindent\textit{2. ispace, inc., Sumitomo Fudosan Hamacho Building 3F, Tokyo, Japan.}
\begin{abstract}
    \normalsize
    Private lunar missions are faced with the challenge of robust autonomous navigation while operating under stringent constraints on mass, power, and computational resources. This work proposes a motion-field inversion framework that uses optical flow and rangefinder-based depth estimation as a lightweight CPU-based solution for egomotion estimation during lunar descent. We extend classical optical flow formulations by integrating them with depth modeling strategies tailored to the geometry for lunar/planetary approach, descent, and landing—specifically, planar and spherical terrain approximations parameterized by a laser rangefinder. Motion field inversion is performed through a least-squares framework, using sparse optical flow features extracted via the pyramidal Lucas-Kanade algorithm. We verify our approach using synthetically generated lunar images over the challenging terrain of the lunar south pole, using CPU budgets compatible with small lunar landers. The results demonstrate accurate velocity estimation from approach to landing, with sub-10\% error for complex terrain and on the order of 1\% for more typical terrain, as well as performances suitable for real-time applications. This framework shows promise for enabling robust, lightweight onboard navigation for small lunar missions.
\end{abstract}
\clearpage

\section{Introduction} \label{sec:introduction}

Autonomous landing is a key enabling technology for sustained lunar exploration. Recent missions by national space agencies, such as ISRO’s \textit{Chandrayaan} series \cite{isroMissionsAccomplishedCharandrayaan}, CNSA’s \textit{Chang’e} program \cite{cnsaMilestonesMarkchangE}, and NASA’s \textit{Artemis} initiative \cite{nasaArtemisNASA}, along with the Lunar Gateway collaboration \cite{nasaGatewayNASA}, have renewed global interest in returning to the Moon. In parallel, private companies including ispace (HAKUTO-R) \cite{ispaceincHAKUTORMissions}, Firefly Aerospace \cite{fireflyspaceBlueGhost}, and Intuitive Machines \cite{intuitivemachinesIM1Intuitive} have begun developing their own commercial lunar landers, following the momentum created by the Google Lunar X Prize \cite{google2007}. Despite recent progress, several failed landing attempts \cite{ispace2023m1, ispaceincHAKUTORMissions, rannard2025} highlight the ongoing challenge of achieving fully autonomous, robust, and cost‑efficient lunar descent.

A major challenge for lunar and planetary landings is the absence of a global navigation satellite system (GNSS). During descent, portions of the trajectory may be out of contact with ground control, requiring the spacecraft to estimate its state autonomously. Relying solely on IMU data leads to drift accumulation, threatening navigational accuracy and mission success \cite{chelmins2009kalman, jaskot2010inertial}. Robust, self-contained egomotion estimation is therefore essential.

State‑of‑the‑art navigation systems \cite{epp2007autonomous,frapard2006vision,verweld2013relative,xu2020navigation,li2006autonomous,li2007vision,johnson2008overview} typically combine active sensors such as LiDAR, laser rangefinders, and radar altimeters with monocular or stereo vision and IMUs, often integrated through Kalman filtering and terrain-relative navigation (TRN). While these methods achieve high precision, their mass and power consumption make them unsuitable for small private landers, underscoring the need for lightweight and low‑power alternatives.

In contrast, methods from the unmanned aerial vehicle (UAV) domain demonstrate that vision‑based optical‑flow techniques can achieve efficient, accurate egomotion estimation using standard optical-flow estimation for tracking image features, approximating surfaces as planar/near-planar, and utilizing homography to estimate velocity, therefore using only a monocular camera and IMU \cite{grabe2012robust,grabe2015nonlinear,panahandeh2013vision,wang2013vision}. Extending such passive systems to lunar descent requires robust sensor pairing. Vision‑only approaches like DLR’s Crater Navigation system (CNAV) \cite{maass2020crater} achieve high accuracy via crater matching but depend on known terrain and lighting. LiDAR‑based systems, such as NDL and HDL \cite{xu2020navigation,epp2007autonomous}, offer strong autonomy but are heavy, high‑power, and limited in range. Laser rangefinders (LRF) offer a balanced compromise—lightweight, fast, and precise—providing scale information when paired with a vision system \cite{williams2022next,miso1999optical}.

This work introduces a motion‑field inversion framework that combines sparse Lucas–Kanade optical flow from a monocular camera with planar and spherical depth models parameterized by rangefinder data and IMU attitude estimates. The framework is evaluated across orbital and terminal descent regimes over diverse lunar terrains, under computational constraints consistent with small lunar landers.

\begin{figure}
    \centering
    \subfigure{
        \includegraphics[width=0.48\linewidth]{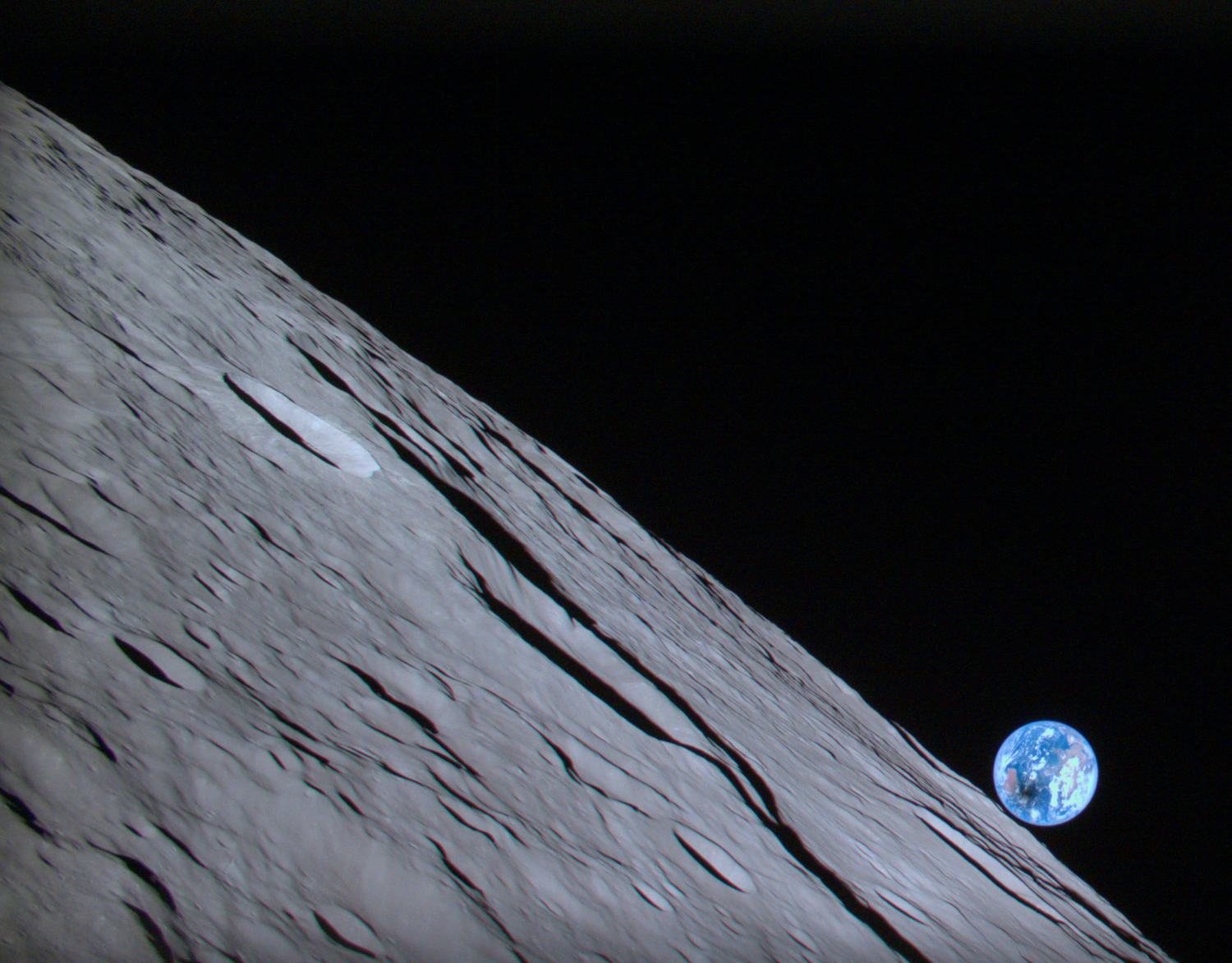}
    }
    \hfill
    \subfigure{
    \centering
    \includegraphics[width=0.48\linewidth]{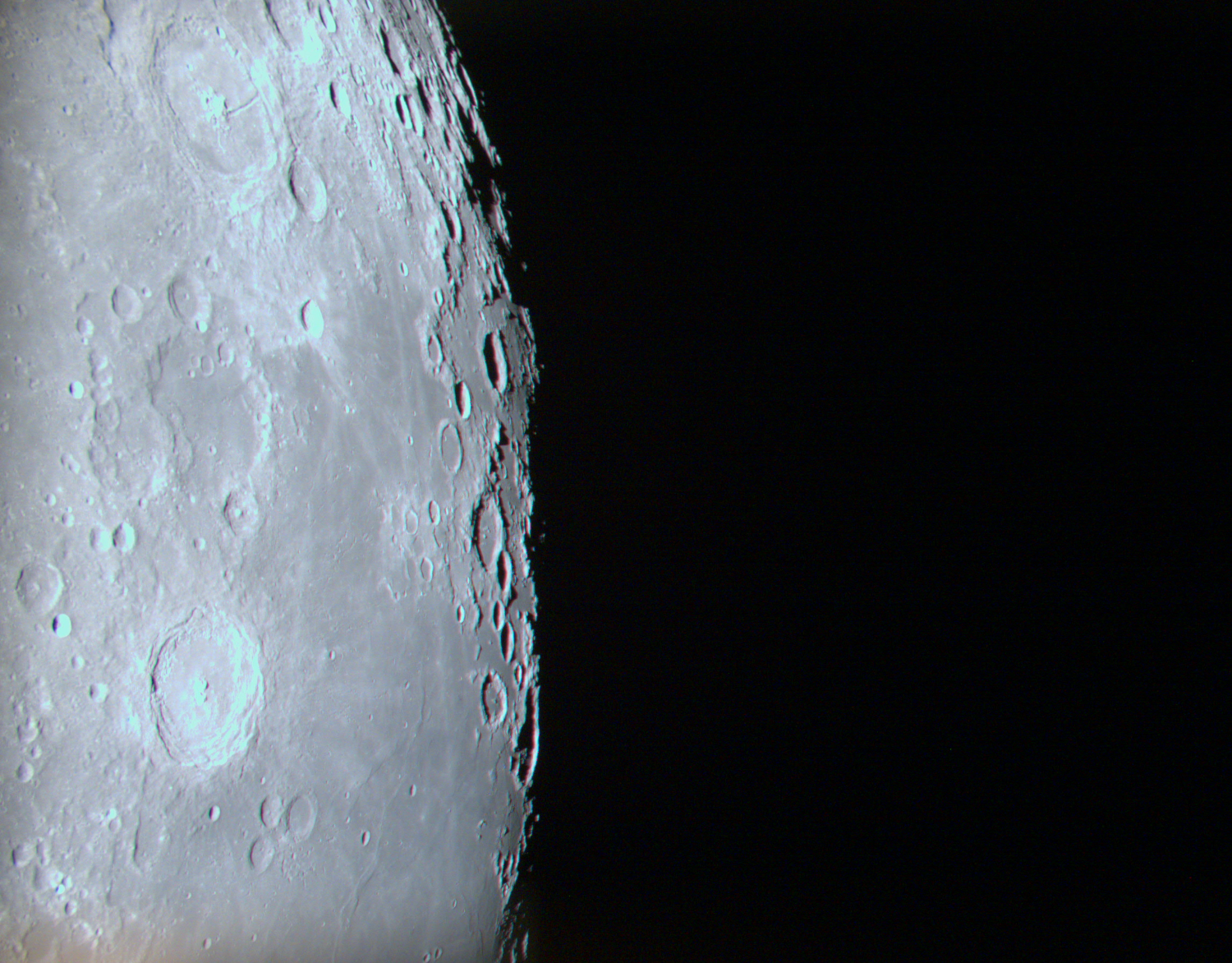}
    }
    \caption{Lunar horizon images captured by ispace's Hakuto‑R lander during Mission 1, April 2023 \cite{ispace2023news, planetary2023hakuto}. ispace provided some insights on the typical hardware constraints and trajectory characteristics of a small private lunar mission.}
\end{figure}

\section{Optical Flow-based Navigation}
\label{sec:methodology}

This section outlines the methodology used for optical flow-based navigation, which leverages image sequences to estimate egomotion in a visually static environment such as the lunar surface.

\subsection{Motion Field and Egomotion Estimation} \label{subsec:MFE-egomotion}

The motion field describes the apparent motion of scene points across the image plane, resulting from the relative movement between a camera (observer) and its surrounding environment. It is defined as the 2D projection of 3D motion onto the image plane, typically modeled using the pinhole camera approximation. This is visualized in \cref{fig:plane}. Under the assumption of a static environment, which is well-justified for the inert and lifeless lunar surface, the observed motion field is directly attributable to the egomotion of the camera, enabling the derivation of a deterministic relationship between the two.

To express such relationship we first define the camera reference frame, shown in \cref{fig:plane}, as $\mathscr{F}^c = [\hat{x}, \hat{y}, \hat{z}]^T$ with origin in $\mathbf{O}$. For compactness, it is henceforth assumed that vectors are expressed in the camera reference frame unless otherwise specified. For a scene point with image coordinates $(\tilde{x}, \tilde{y})$ and corresponding optical flow $\mathbf{u}=[u, v]^T$ we have \cite{gupta19953}:

\begin{equation}
    \begin{cases}
    u = (\tilde{x} v_{ z} - f_x v_{x}) d(\tilde{x},\tilde{y}) - f_x q + r\tilde{y} + \frac{p\tilde{x}\tilde{y}}{f_y} - \frac{q\tilde{x}^2}{f_x} \\
    v = (\tilde{y} v_{z} - f_y v_{y}) d(\tilde{x},\tilde{y}) + f_y p - r\tilde{x} - \frac{q\tilde{x}\tilde{y}}{f_x} + \frac{p\tilde{y}^2}{f_y}
    \end{cases}
    \label{eq:egomotion}
\end{equation}
Alternatively, in matrix form we have:
\begin{equation}
 \mathbf{u} = d(\tilde{x},\tilde{y})
\underbrace{\begin{bmatrix}
    -f_x & 0 & \tilde{x} \\
    0 & -f_y & \tilde{y} 
\end{bmatrix}}_{\mathbf{L}_{t(\tilde{x},\tilde{y})}}
\mathbf{v}
+
\underbrace{\begin{bmatrix}
    \frac{\tilde{x}\tilde{y}}{f_y} & -\left(f_x+\frac{\tilde{x}^2}{f_x}\right) & \tilde{y} \\
    f_y + \frac{\tilde{y}^2}{f_y} & -\frac{\tilde{x}\tilde{y}}{f_x} & -\tilde{x}
\end{bmatrix}}_{\mathbf{L}_{\omega(\tilde{x},\tilde{y})}}
\mb{\omega},
\end{equation}
where the linear egomotion has components $\mathbf{v}= [v_{x}, v_{y}, v_{z}]^T$ and angular egomotion has $\mb{\omega} = [p, q, r]^T$; $\mathbf{L}_t$ and $\mathbf{L}_\omega$ are the translational and rotational contributions, respectively, that map from image velocity to translation and rotation in the camera reference frame; $f_x$ and $f_y$ denote the vertical and horizontal focal lengths of the camera, and the function $d(\tilde{x},\tilde{y})$ is the inverse depth-map.

With access to a depth map and knowledge of attitude and angular velocities, \cref{eq:egomotion} allows for an unambiguous estimation of the vehicle's egomotion \cite{horn1988motion}, demonstrated later in \cref{subsec:motion-field-inversion}. In the absence of direct depth measurements, approximate terrain geometry can be inferred through a rangefinder and an assumed depth model. The formulation of such a depth map is a critical step, discussed in the subsequent section.

\subsection{Depth Models}\label{subsec:depth-models}
The depth map plays a crucial role as the bridge between the observed image-plane motion (optical flow) and real-world motion (egomotion). On the Moon, absolute depth maps are difficult and computationally costly to acquire in real-time \cite{ye2020lunar, wang2022improvement, jia2018dispnet}. Thus, incorporating simple depth models in conjunction with a lightweight rangefinder measurement enables approximation of the depth map, and with that the estimation of the egomotion from the optical flow. 

In this paper, two main depth models have been explored: (1) A planar depth model, and (2) a spherical depth model, illustrated in \cref{fig:depth_map_geometry}. A fixed planar depth model has been demonstrated as a surprisingly lightweight but effective model for UAV egomotion estimation \cite{grabe2012robust, grabe2012board, strydom2014visual}. As an extension of the fixed planar model, to approach more complex, sloping terrains, a slope estimated planar model is developed. At higher altitudes, a planar description of the lunar surface becomes less accurate, while a spherical depth model becomes a better approximation.

\subsubsection{Plane Approximation}\label{subsubsec:plane-approx}

\begin{figure}
    \centering
    \subfigure[Planar depth model]{
        \includegraphics[width=0.47\linewidth]{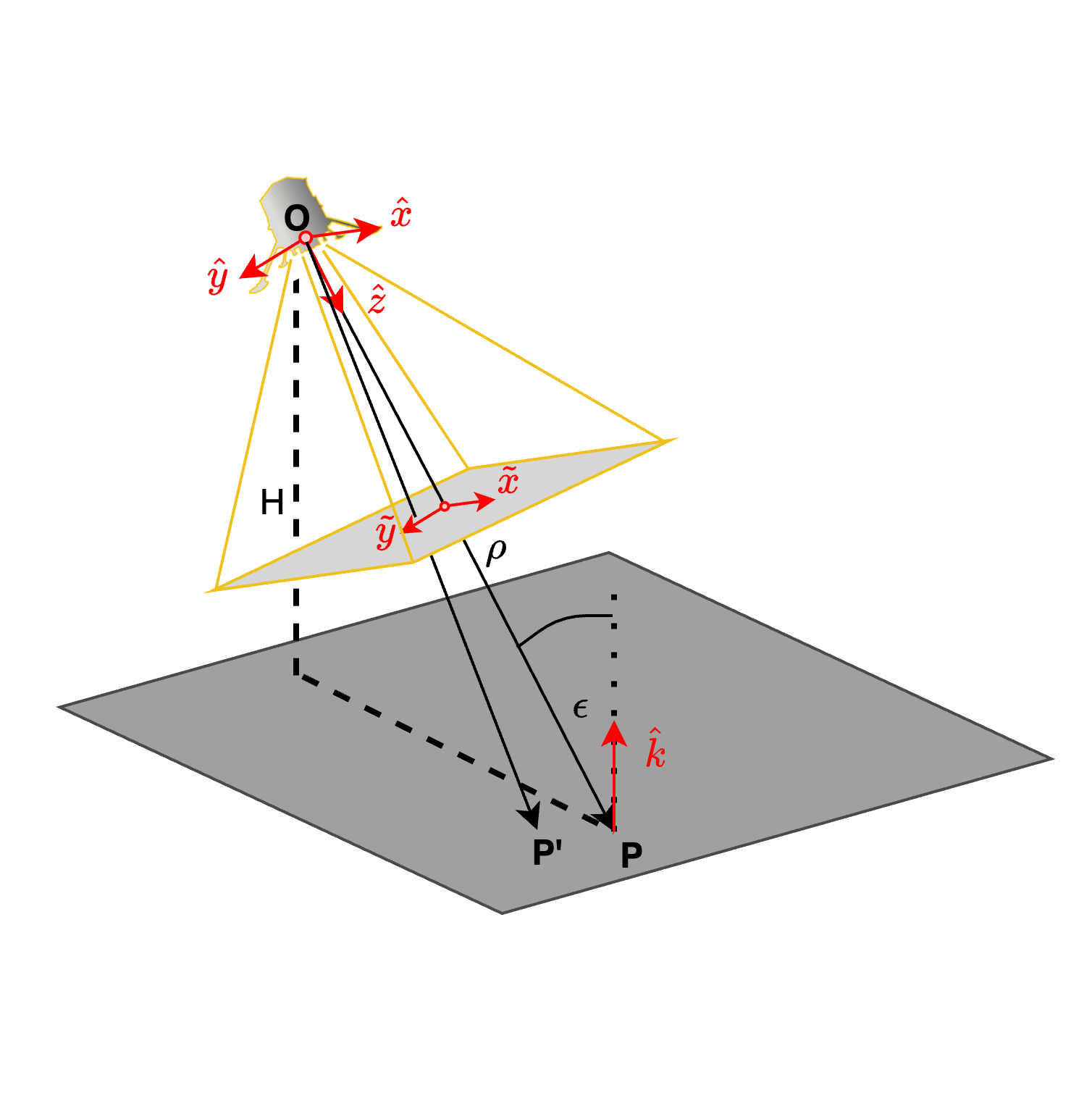}
        \label{fig:plane}
    }
    \hfill
    \subfigure[Spherical depth model]{
        \includegraphics[width=0.47\linewidth]{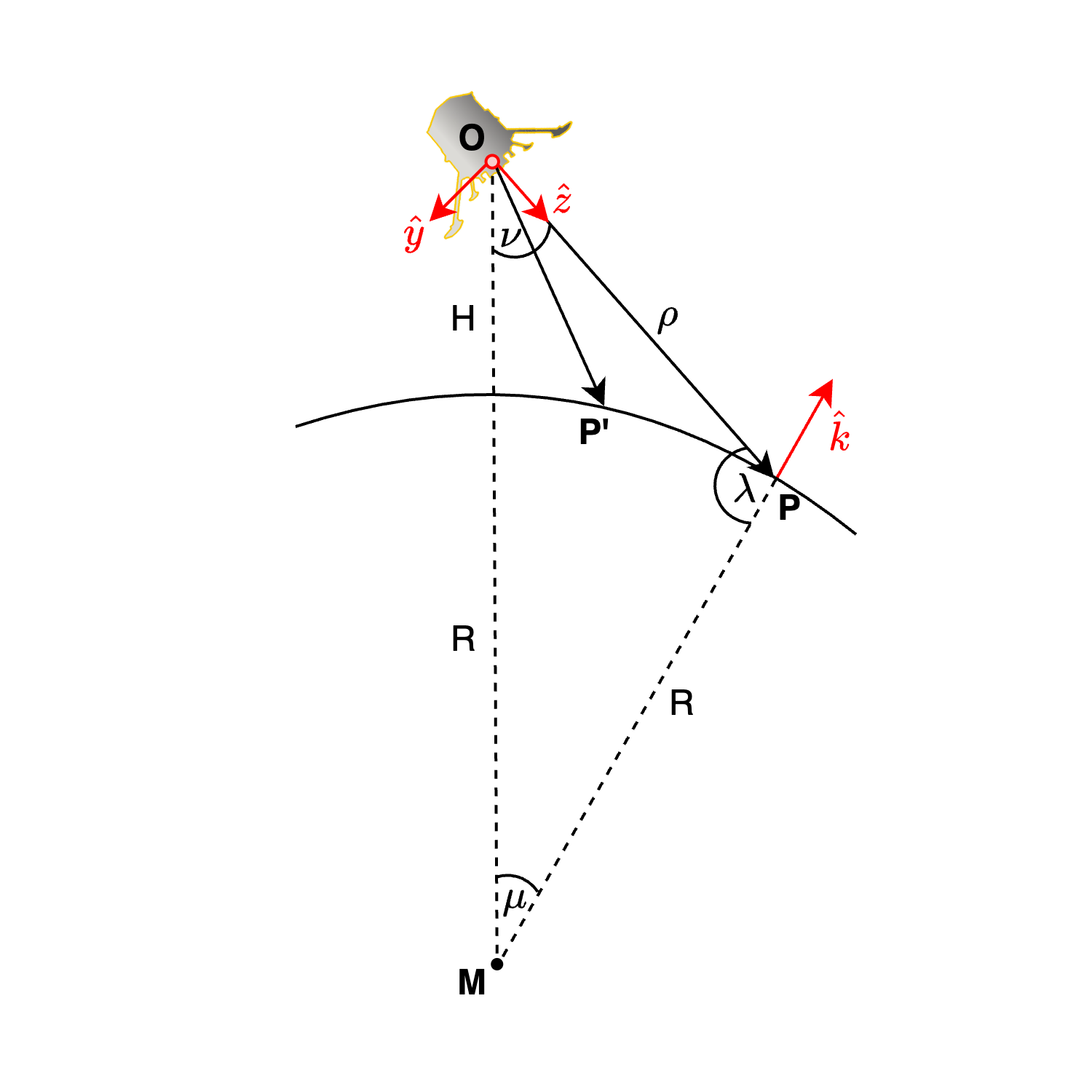}
        \label{fig:sphere}
    }
    \caption{Illustrations of the two principal depth map geometries evaluated in this work. 
        The planar depth model assumes the local lunar surface can be approximated by a plane and is suitable for low-altitude operations, with optional slope estimation to handle inclined terrain. 
        The spherical depth model represents the lunar surface as part of a sphere and offers improved accuracy at higher altitudes where curvature effects become significant.}
    \label{fig:depth_map_geometry}
\end{figure}

If the local lunar surface is assumed to be a plane, then one can derive an expression for the depth map using the geometry in \cref{fig:plane}. To start, take the vectors $\mathbf{r}_{OP} = \left[X, Y, Z\right]^T$ (where subscript $OP$ indicates a vector from \textbf{O} to \textbf{P}) and $\mathbf{\hat{k}} = \left[    \alpha, \beta, \gamma \right]^T$. A key insight is that the dot product of these two vectors gives us H and subsequently relates $\alpha$, $\beta$, and $\gamma$ to the depth $Z$:

\begin{align}
    \mathbf{r}_{OP} \cdot \mathbf{\hat{k}} &= H \\
    \alpha X + \beta Y + \gamma Z &= H \\
    \frac{\alpha}{Z} X + \frac{\beta}{Z} Y + \gamma &= \frac{H}{Z}
\end{align}

Then, we define $\tilde{x} = \frac{X}{Z}$ and $\tilde{y} = \frac{Y}{Z}$ to convert to the image coordinates mentioned in \cref{subsec:MFE-egomotion}. To further simplify, we define $\bar{\alpha} = \frac{\alpha}{H}$, $\bar{\beta} = \frac{\beta}{H}$, and $\bar{\gamma} = \frac{\gamma}{H}$ and we get:

\begin{equation} \label{eq:baralpha}
        \bar{\alpha} \tilde{x} + \bar{\beta} \tilde{y} + \bar{\gamma} = \frac{1}{Z} = d(\tilde{x}, \tilde{y})
\end{equation}

The goal is to express the depth equation in terms of the known range-to-ground \(\rho\), rather than the altitude \(H\). From geometry, we know that the component of the unit surface normal \(\mathbf{\hat{k}}\) along the optical axis is
\begin{equation}
    \gamma = \cos \epsilon,
    \label{eq:gamma-epsilon}
\end{equation}
where $\epsilon$ is the angle between $\mathbf{\hat{k}}$ and the line-of-sight vector ($\vec{r}_{PO}$). The relationship between $\rho$ and $H$ is then given by
\begin{equation}\label{eq:rho-epsilon}
        H = \rho \cos \epsilon.
\end{equation}

Combining \cref{eq:gamma-epsilon} and \cref{eq:rho-epsilon} by substituting $\cos \epsilon$ gives us $\bar{\gamma} = \frac{1}{\rho}$ and then substituting into \cref{eq:baralpha}.

\begin{equation} \label{eq:planar-depth}
    \bar{\alpha} \tilde{x} + \bar{\beta} \tilde{y} + \frac{1}{\rho} = \frac{1}{Z} = d(\tilde{x}, \tilde{y})
\end{equation}
yielding a depth function parameterized directly by \(\rho\), a measurable quantity during descent.

For more adaptability of this plane approximation of the surface — coined the \textit{fixed} planar model, instead of assuming $\alpha$ and $\beta$ are fixed, one can also estimate these parameters during the motion field inversion. This can be referred to as the \textit{slope estimation} planar model, as was mentioned in \cref{subsec:MFE-egomotion}.


\subsubsection{Sphere Approximation} \label{subsubsec:sphere-approx}

If the surface is modeled by a sphere, which becomes an increasingly accurate description at higher altitudes, a different expression for the depthmap can be derived. The geometry is used as shown in \cref{fig:sphere}. Given $\mathbf{r}_{OP} = \begin{bmatrix}
    0, 0, \rho
\end{bmatrix}^T$ and $\mathbf{r}_{PM} = \begin{bmatrix}
    -R \alpha, -R \beta, R \gamma
\end{bmatrix}^T$, with vector addition, one gets $\mathbf{r}_{OM} = \begin{bmatrix}
    -R \alpha, -R \beta, \rho - R \gamma
\end{bmatrix}^T$. Next, points need to be described in the field of view around this point. For this, point $\mathbf{P'}$ is defined, from which an expression for $Z$ can be developed using $\mathbf{r}_{MP'} = \mathbf{r}_{OP'} - \mathbf{r}_{OM}$. Their magnitude is equivalent, which leads to:

\begin{equation} \label{eq:z-rho-sphere}
    R = || \begin{bmatrix}
    X \\ Y \\ Z
    \end{bmatrix} - \begin{bmatrix}
    -R\alpha \\ -R\beta \\ \rho - R\gamma
    \end{bmatrix} ||
\end{equation}
\cref{eq:z-rho-sphere} gives a quadratic equation in Z, which can be solved. To break down the full expression of the spherical depth map, a number of intermediate expressions are defined in \cref{eq:a,eq:b,eq:c,eq:d}. These expressions are analogous to the terms from the well-known abc-formula to solve a quadratic equation.
\begin{align}
    a &= \tilde{x}^2 + x_y^2 + 1 \label{eq:a} \\
    b &= 2 \tilde{x} R \alpha + 2 \tilde{x} R \beta + 2 R \gamma - 2 \rho \label{eq:b}\\
    c &= R^2 \alpha^2 +R^2\beta^2 + \rho^2 + R^2\gamma^2 - 2R\rho\gamma - R^2 \label{eq:c} \\
    d &= b^2 - 4 a c \label{eq:d}
\end{align}








Finally, \cref{eq:a,eq:b,eq:c,eq:d} can be combined to form an expression for the depthmap in \cref{eq:spherical-depthmap}
\begin{equation} \label{eq:spherical-depthmap}
    d(\tilde{x}, \tilde{y}) = \frac{1}{z(\tilde{x}, \tilde{y})} = \frac{1}{\frac{-b - \sqrt{d}}{2a}}
\end{equation}

\cref{eq:spherical-depthmap} is the spherical equivalent of \cref{eq:planar-depth}; we can verify that with $R \to +\infty$ the spherical depth map reduces to the planar one. 

\subsection{Optical Flow Estimation} \label{subsec:OF-est}

The motion field's components $u$ and $v$ in \cref{eq:egomotion} can be approximated by computing the optical flow. A dense optical flow is not needed, and it's rather preferred to have fewer well-defined and easy to track features, that results in a sparse but precise optical flow prediction. Considering also the need for real-time computation onboard, the iterative Lucas-Kanade method with pyramids was identified as a low computation cost, robust optical flow estimator with subpixel accuracy \cite{lucas1981iterative,bouguet2001pyramidal,shi1994good}.

\begin{figure}
    \centering
    \includegraphics[width=\linewidth]{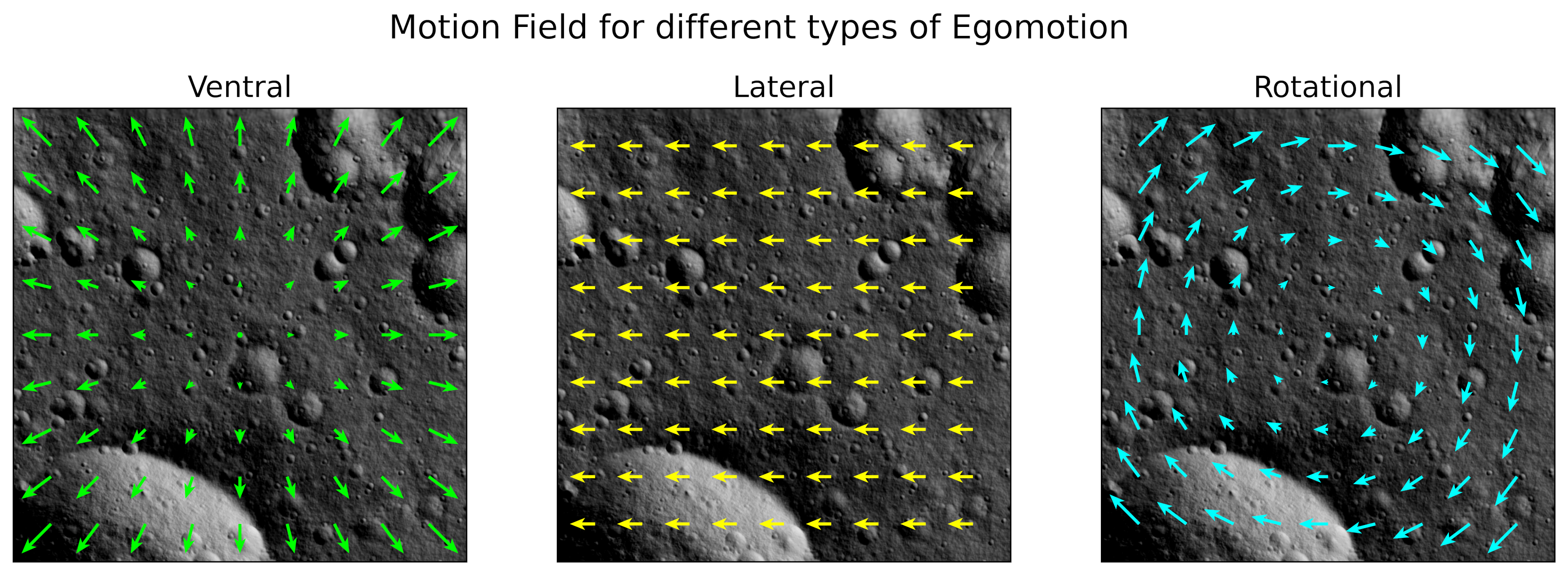}
    \caption{Example of Motion Fields observed when looking ventrally for different regimes of Egomotion. Ventral corresponds to egomotion ventrally toward the surface. Lateral corresponds to egomotion parallel to the surface. Rotational corresponds to rotation about the line-of-sight of the camera.}
    \label{fig:optical-flow}
\end{figure}

\subsection{Motion Field Inversion -- Egomotion Estimation} \label{subsec:motion-field-inversion}

\begin{figure}
    \centering
    \includegraphics[width=\linewidth]{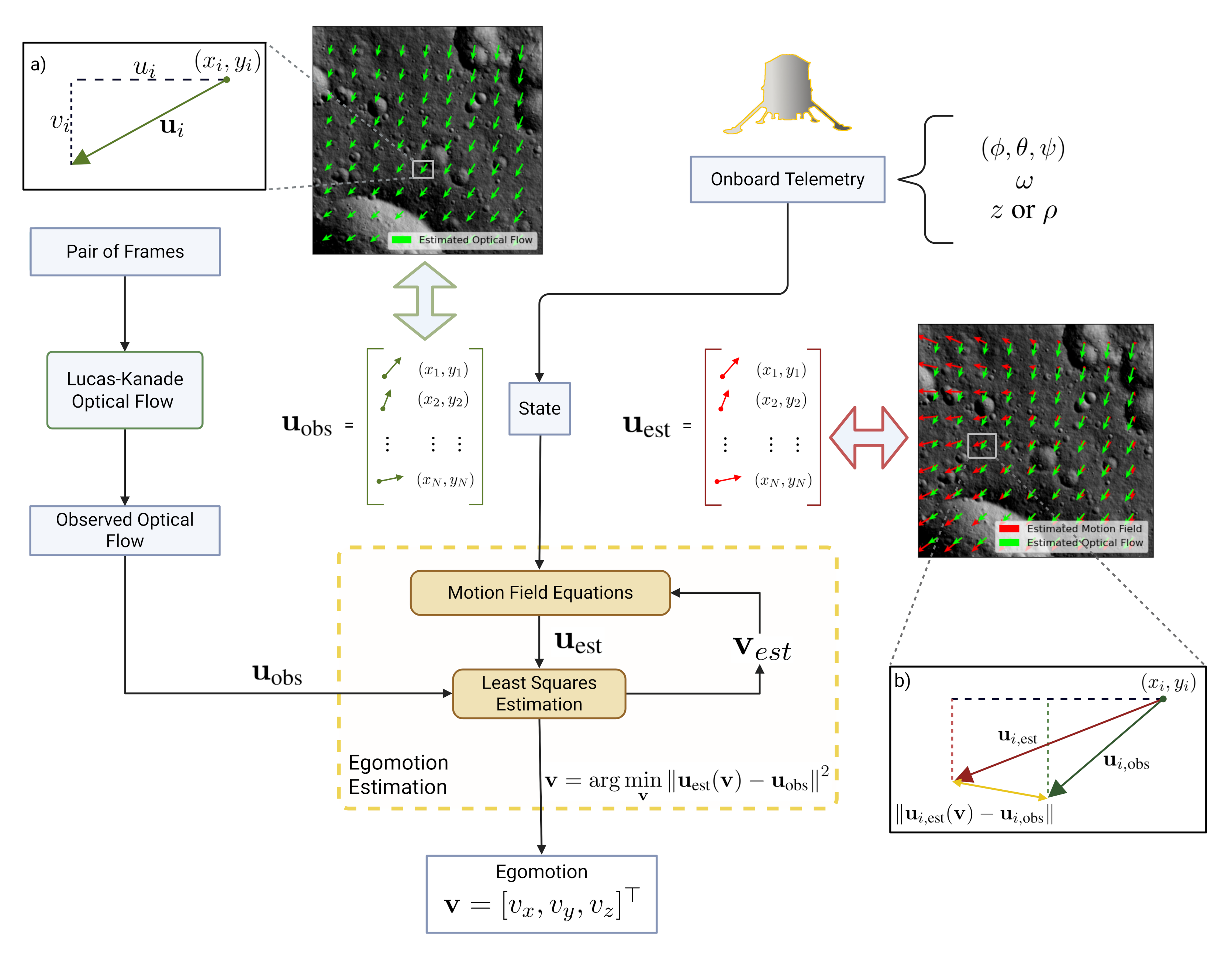}
    \caption{Flowchart describing the full motion field inversion from frames and onboard telemetry to egomotion estimation. a) Optical flow arrow components. b) Residual calculation between observed optical flow and estimated motion field from least squares estimated velocity, $\mathbf{v}_{\text{est}}$. Note: For a linear system of equations (i.e. when using flat planar or spherical depth models), the iterative residual least squares calculation segment is not required. However, when a non-linear system of equations is present (i.e. when estimating further parameters such as slope estimation planar model or attitude estimation) iterative non-linear least squares is required - this is shown in the above diagram. }
    \label{fig:motion-field-inversion}
\end{figure}

The components introduced so far -- motion field equations, approximate depth models, and sparse optical flow estimation -- can now be combined into a unified framework for estimating egomotion using only onboard imagery, a rangefinder measurement, and IMU-derived angular velocities.

The system described in \cref{eq:egomotion} provides two equations for each point at which the optical flow is tracked. In the simplest cases (i.e. flat planar or sphere approximation), there are three unknowns corresponding to translational velocities $v_x$, $v_y$, and $v_z$.  Consequently, only two unique, tracked points are sufficient to form an overdetermined system and subsequently solve the motion field equations. In the additional case of the slope estimation planar model, the relative inclination between the camera and the lunar surface is also estimated, represented by the angles $\alpha$ and $\beta$, thereby extending the number of unknowns to five and therefore the minimum number of unique, tracked points to three. To leverage the typically higher number of tracked points provided by the Lucas-Kanade algorithm, and to mitigate the influence of noise in individual measurements, a least-squares estimation approach is adopted. 

The observed motion field, given a stationary environment and moving observer, is described precisely by \cref{eq:egomotion}. Given a pair of frames, an optical flow estimator tracks a number of features between those frames. Each tracked feature $i$ at normalized image coordinates $(\tilde{x}^{(i)}, \tilde{y}^{(i)})$ contributes a pair of equations:

\begin{equation}
    \underbrace{\begin{bmatrix} u^{(i)}\\v^{(i)}    \end{bmatrix}}_{\mathbf{u}^{c(i)}} = d^{(i)}\mathbf{L}_t^{(i)}\mathbf{v}+\mathbf{L}_\omega^{(i)}\mb{\omega},
\end{equation}

Where abbreviations $d^{(i)}=d^{(i)}(\tilde{x},\tilde{y})$, $\mathbf{L}_t^{(i)} = \mathbf{L}_t^{(i)}(\tilde{x},\tilde{y})$, $\mathbf{L}_w^{(i)} = \mathbf{L}_w^{(i)}(\tilde{x},\tilde{y})$ have been used. Stacking the constraints from $N$ such features yields a system of equations:

\begin{equation}
\underbrace{
\begin{bmatrix}
\mathbf{u}^{(1)} \\
\vdots \\
\mathbf{u}^{(N)} 
\end{bmatrix}
}_{\mathbf{U} \in \mathbb{R}^{2N}}
=
\underbrace{
\begin{bmatrix}
 d^{(1)} \mathbf{\mathbf{L}_t}^{(1)} \\
\vdots \\
 d^{(N)} \mathbf{\mathbf{L}_t}^{(N)}
\end{bmatrix}
}_{\mathbf{A} \in \mathbb{R}^{2N \times 3}}
\mathbf{v}
+ \underbrace{
\begin{bmatrix}
 \mathbf{L}_\omega^{(1)} \\
\vdots \\
 \mathbf{L}_\omega^{(N)}
\end{bmatrix}
}_{\mathbf{B} \in \mathbb{R}^{2N\times 3}}
\mb{\omega}.
\end{equation}

In compact form, we then have:

\begin{equation}\label{eq:compact-motion-inv}
\mathbf{U} =
 \mathbf{A} \mathbf{v} + \mathbf{B}\mb{\omega}, \quad \text{with} \quad
\mathbf{U} \in \mathbb{R}^{2N}, \;
\mathbf{A} \in \mathbb{R}^{2N \times 3}, \;
\mathbf{v} \in \mathbb{R}^3, \; 
\mathbf{B}\in \mathbb{R}^{2N\times 3}\;
\mb{\omega}\in \mathbb{R}^3
\end{equation}
Each row of $\mathbf{A}$ corresponds to the translation component of a single flow constraint scaled by the inverse depth map $d(\tilde{x}^{(i)}, \tilde{y}^{(i)})$. This can be rearranged to form the standard formulation of a linear system of equations:
\begin{equation} 
\mathbf{A}\mathbf{v}=\mathbf{C}, \quad \text{with} \quad \mathbf{A} \in \mathbb{R}^{2N \times 3}, \mathbf{v} \in \mathbb{R}^3, \mathbf{C}=\mathbf{U}-\mathbf{B}\bm{\omega}\in \mathbb{R}^{2N}
\end{equation}

A solution of the least squares problem $\arg\min_{\mathbf{v}} \left\| \mathbf{A}\mathbf{v}-\mathbf{C} \right\|^2$ can be obtained in closed form using the normal equations: 
$$ \mathbf{v}=(\mathbf{A}^T \mathbf{A})^{-1}\mathbf{A}^T \mathbf{C}$$

This requires that $\mathbf{A}$ has full column rank so that $\mathbf{A}^\top \mathbf{A}$ is invertible. If $\mathbf{Av}=\mathbf{C}$ is exactly consistent, which is rare in real scenarios, this formula yields an exact solution; otherwise, it returns the least-squares solution which minimizes the residual norm $\left\| \mathbf{A}\mathbf{v}-\mathbf{C} \right\|^2$ use iterative singular value decomposition methods \cite{anton2005elementary}.

In the slope estimation case, the depth map also depends on $\alpha \text{ and } \beta$, which leads to a non-linear system of equations — as can be seen by substituting the new depth map $d(\tilde{x}^{(i)}, \tilde{y}^{(i)},\alpha,\beta)$ into \cref{eq:egomotion}. Consequently, a solution can be obtained by iterative non-linear residual least squares: 
\begin{equation}
    \mathbf{v}=\arg\min_{\mathbf{v}} \left\| \mathbf{u}_{\text{est}}(\mathbf{v}) - \mathbf{u}_{\text{obs}} \right\|^2 
\end{equation}
where $\mathbf{u}_{\text{obs}}\in \mathbb{R}^{2N}$ is the stacked observed optical flow from an optical flow estimator,  $\mathbf{u}_{\text{est}}(\mathbf{v})\in \mathbb{R}^{2N}$ is the stacked flow vector predicted based upon an estimated translational velocity, depth model and known state information ($(\phi, \theta, \psi)$, $[p, q, r]^T$, and $z$ or $\rho$). The solution here is iteratively minimized using the trust region reflective optimization algorithm \cite{2020SciPy-NMeth}.

To summarize, the flow of this process for a pair of frames — illustrated in \cref{fig:motion-field-inversion} — is the following:

\begin{enumerate}
    \item Optical Flow Estimation -- Estimate optical flow for a pair of frames using Lucas-Kanade.
    \item Parameter initialization -- Start with an initial guess for egomotion and depth model parameters.
    \item Motion Field Computation -- Compute the motion field from the current parameters taking into account the known attitude angular velocity and depth model.
    \item Residual Calculation -- Compare predicted motion field to observed optical flow.
    \item Parameter update -- Adjust parameters to reduce the residual using iterative algorithm.
    \item Repeat until convergence.
\end{enumerate}

To properly demonstrate the potential and limitations of this method alone, techniques such as filtering are deliberately omitted. The solutions for each pair of frames are estimated independently, with the sole exception of using predictions from the previous state as the initial guess for the subsequent one.

\begin{figure}
    \centering
    \includegraphics[width=\linewidth]{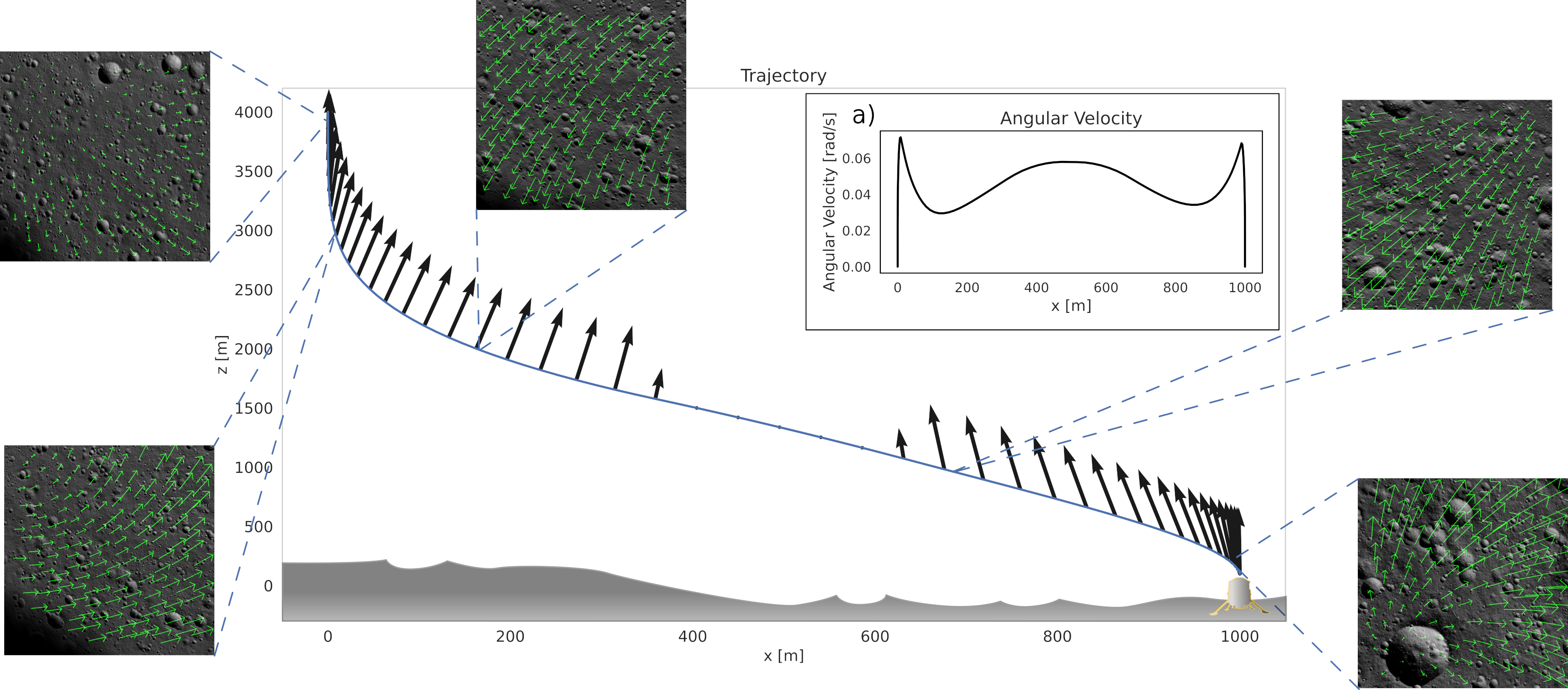}
    \caption{Example Landing Sequence trajectory showing observed sparse optical flow at various frames throughout the trajectory. a) Shows the non-zero magnitude of the angular velocity throughout the whole trajectory -- explaining the curl of the observed optical flow fields. }
    \label{fig:landing-trajectory-composite}
\end{figure}

\section{Experimental Details}
\label{sec:experimental_details}

\subsection{Trajectory Generation and Simulation}

We simulated realistic lunar descent trajectories using a previously developed six‑degrees‑of‑freedom lander model \cite{azzalini2023generation}, equipped with a primary descent engine and attitude control via four secondary thrusters. Spacecraft parameters, including mass properties, thrust capability, and moments of inertia, were selected to be representative of the Hakuto mission profile \cite{ispaceincHAKUTORMissions}. Based on these dynamics, we formulated a minimum‑fuel optimal control problem and discretized it using the Hermite–Simpson direct transcription method. The problem was implemented using \texttt{AMPL} to obtain a differentiable nonlinear programming formulation, and solved via sequential quadratic programming.
The optimized descent trajectories served as inputs to \texttt{PANGU} (Planet and Asteroid Natural Scene Generation Utility) \cite{martin2021pangu}, a validated planetary rendering engine \cite{martin2019pangu} capable of producing photorealistic lunar imagery with accurate terrain topography, albedo, and illumination. Each \texttt{PANGU} frame was augmented with a corresponding depth value for the surface point projected at the image center, emulating a laser rangefinder. This synthetic range measurement enables absolute scale recovery in monocular visual odometry, thereby reducing drift typical of pure IMU-based navigation.\cite{zhong2020direct}.

\subsection{Trajectory Types and Landing Sites}
\label{sec:trajectory}
\begin{table}[htbp]
\centering
\renewcommand\cellalign{c} 
\caption{Trajectory details for different scenarios evaluated in this paper. Altitude is measured from the reference spheroid. Note: Final altitude of landing trajectories are all 100m above the desired landing site. Hohmann transfer has no landing site. End-to-end transfer to landing ends directly at a landing site. Velocities defined in the local terrestrial frame}
\begin{tabular}{
    >{\centering\arraybackslash}m{2.2cm}  
    >{\centering\arraybackslash}m{2.5cm}    
    >{\centering\arraybackslash}m{1.2cm}  
    >{\centering\arraybackslash}m{1.2cm}  
    >{\centering\arraybackslash}m{1.5cm}  
    >{\centering\arraybackslash}m{1.5cm}  
    >{\centering\arraybackslash}m{1.5cm}  
    >{\centering\arraybackslash}m{1.5cm}  
}
\toprule
\multicolumn{2}{c}{\multirow{3}{*}{\makecell{Trajectory}}}
& \multicolumn{2}{c}{Altitude (km)} 
& \multicolumn{4}{c}{Velocity (m/s)} \\
\cmidrule(lr){3-4} \cmidrule(lr){5-8}
& & \multirow{2}{*}{\makecell{Initial}} & \multirow{2}{*}{\makecell{Final}} 
& \multicolumn{2}{c}{Initial} & \multicolumn{2}{c}{Final} \\
\cmidrule(lr){5-6} \cmidrule(lr){7-8}
& & & & $\text{vertical}$ & $\text{horizontal}$ & $\text{vertical}$ & $\text{horizontal}$ \\
\midrule
\multirow{4}{*}{\makecell{Landing}} 
    & Flat         & 4   & 0.1 & 100  & 0    & 0   & 0   \\
    & Peak         & 11.2508   & 7.3508 & 100  & 0    & 0   & 0   \\
    & Crater       & 4   & 0.1 & 100  & 0    & 0   & 0   \\
    & Incline      & 8.6614   & 4.7614 & 100  & 0    & 0   & 0   \\
\midrule
Orbital     & Hohmann Transfer        & 300 & 4   & 0.18 & 1489.26 & 0.25 & 1742.40 \\
\midrule
End-to-end  & Full Transfer to Landing & 102.013 & 0   & 0.217 & 1633.50 & 0.01   & 0.45    \\
\bottomrule
\end{tabular}
\label{tab:trajectory-details}
\end{table}

\begin{figure}
    \centering
    \includegraphics[width=1.0\linewidth]{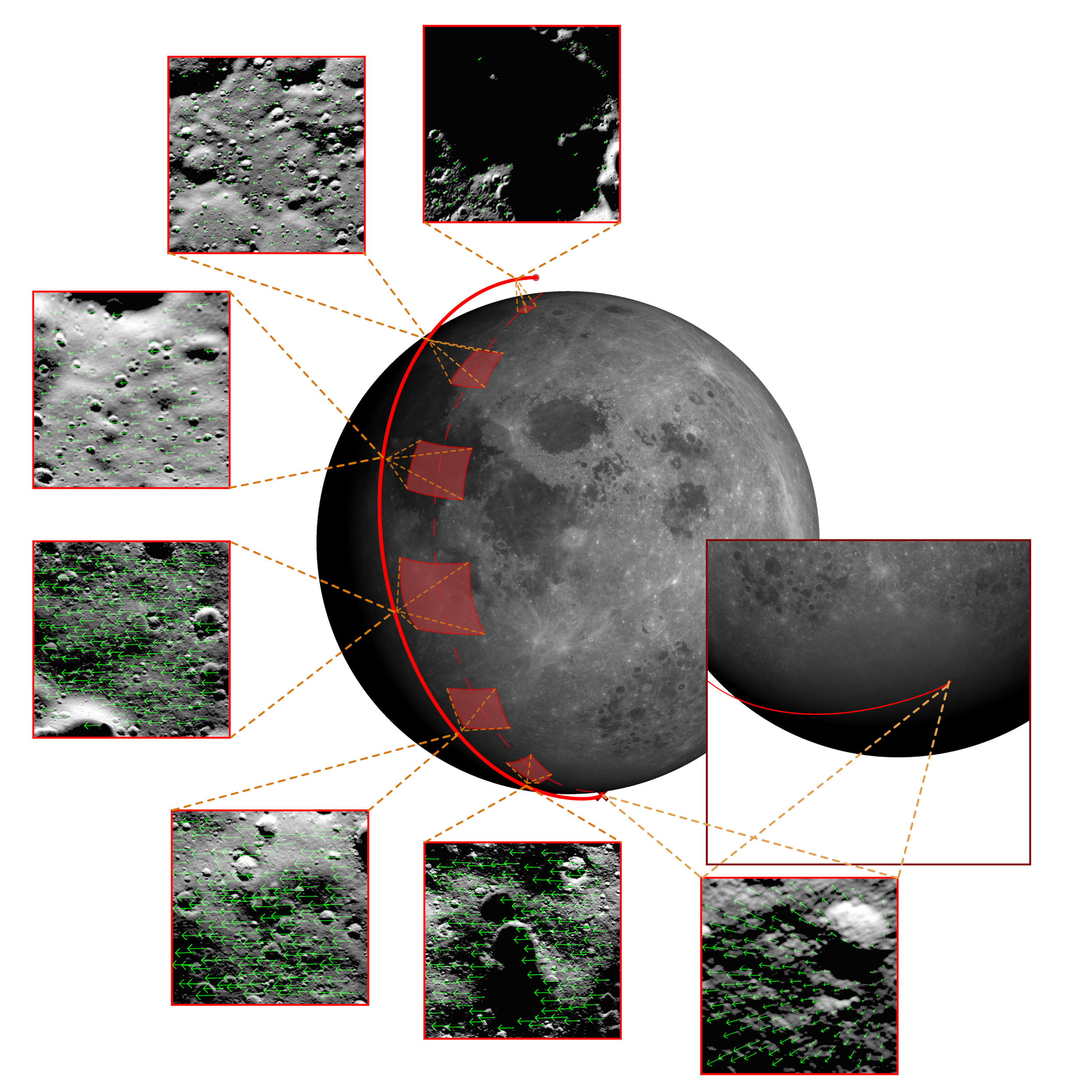}
    \caption{Composite illustration of the end-to-end lunar descent trajectory provided by ispace, combining orbital transfer and landing phases. Frames along the trajectory display the corresponding optical flow fields, showing increasing flow magnitudes as altitude decreases while velocity remains high. Flow intensity diminishes only during the final braking and touchdown phase. The lunar surface texture is included for visualization purposes.}
    \label{fig:ispace-trajectory}
\end{figure}

To evaluate performance under diverse conditions, we generated two classes of trajectories:
\begin{enumerate}
    \item A landing descent profile from 4000\,m to 100\,m depicted in \cref{fig:landing-trajectory-composite}, for which several landing sites were chosen to evaluate robustness to variable terrain.
    \item Hohmann transfers from 300\,km to 4\,km altitude, both around the equator and polar regions.
    \item An end-to-end trajectory,  composed of a polar orbital transfer to landing.
\end{enumerate}
The end-to-end trajectory, illustrated in \cref{fig:ispace-trajectory} was used for empirical verification of the methodology and parameter set across both of the other regimes (landing descent and orbital transfer) simultaneously. 

\begin{figure}
    \centering
    \subfigure[Ventral view]{
        \includegraphics[height=0.33\linewidth]{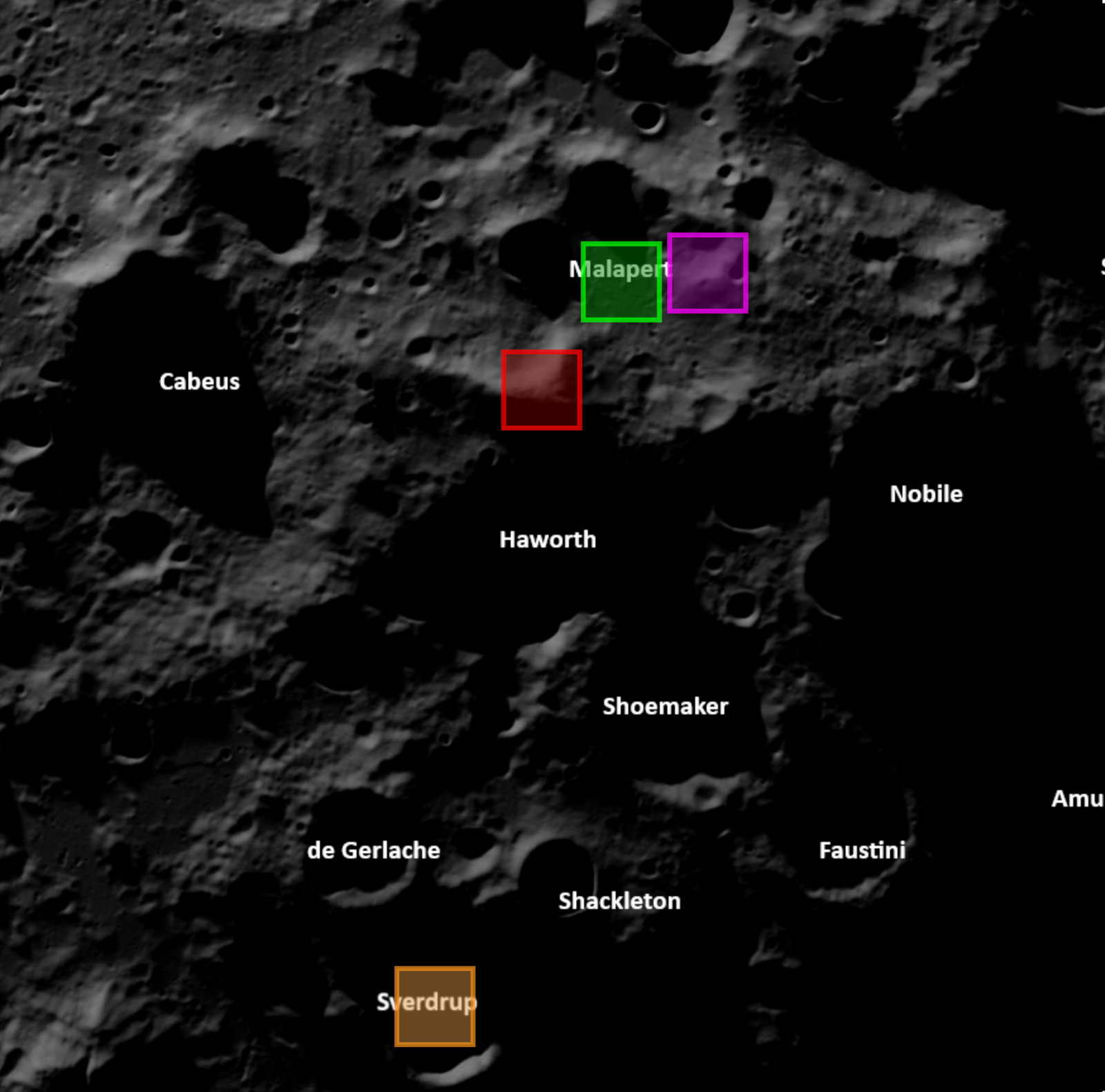}
    }
    \subfigure[Local view]{
        \includegraphics[height=0.33\linewidth]{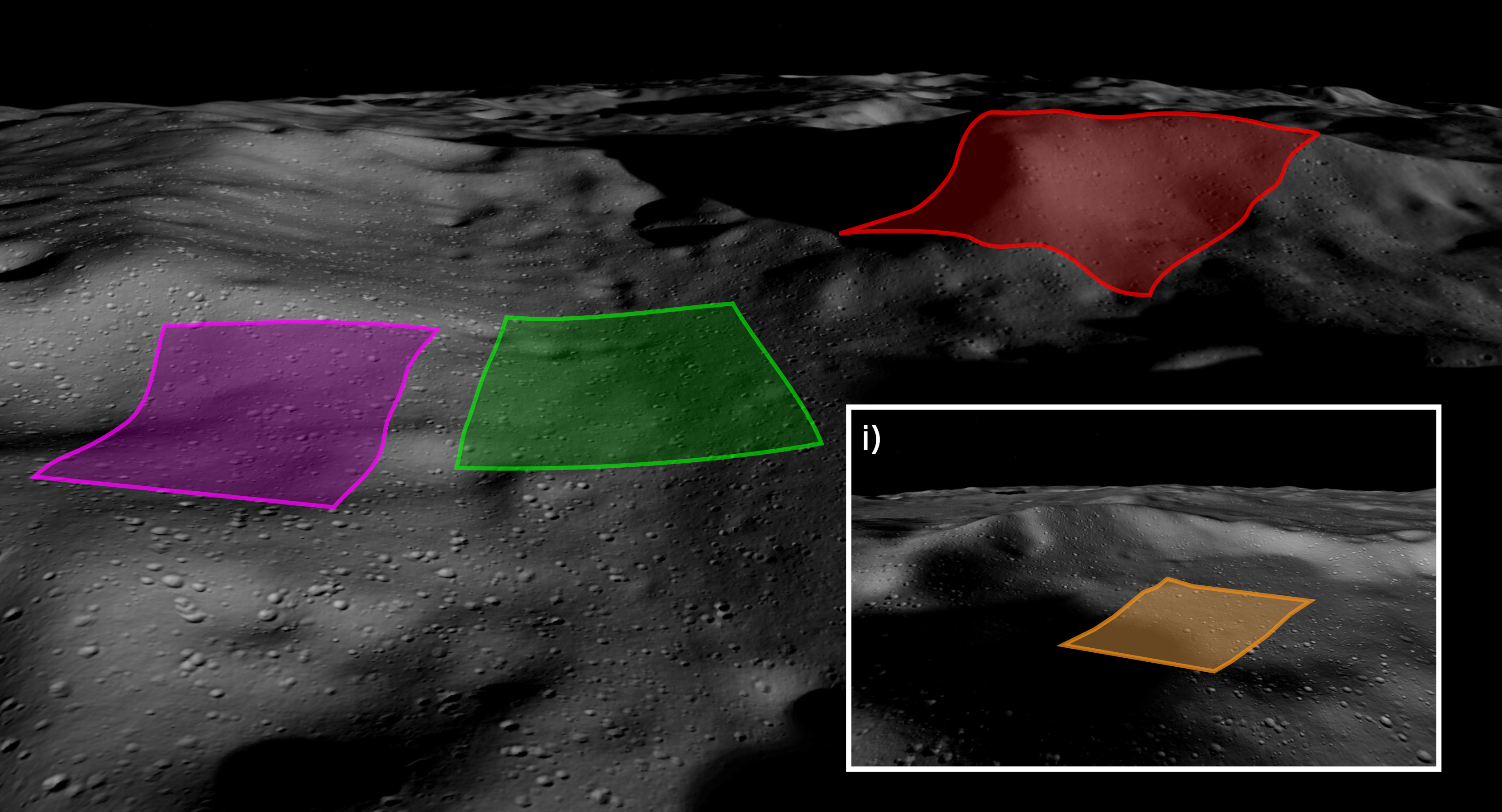}
    }
    \caption{Selected landing sites within the Malapert region used for robustness evaluation. The \textit{Crater} landing site in green, the \textit{Incline} landing site in purple, the \textit{Peak} landing site in red, and the \textit{Flat} landing site in orange. Ventral (a) and local (b) visualizations are shown under realistic illumination conditions.}
    \label{fig:landing_sites}
\end{figure}

For the landing descent trajectory, illustrated in \cref{fig:landing-trajectory-composite}, a number of landing sites were selected to challenge our optical flow-based egomotion estimation method. Three terrains within the Malapert region were prioritized: a relatively flat area surrounded by inclined regions at the crater base (green) henceforth denoted by the \textit{Crater} landing site, an inclined region (purple) denoted by \textit{Incline}, and the Malapert peak (red) denoted by \textit{Peak}, offering increasing levels of geometric complexity. A simpler, arbitrary flat site near the south pole served as a control (orange) denoted by \textit{Flat}. \Cref{fig:landing_sites} illustrates the selected sites with ventral and local visualizations under realistic lighting.

The end‑to‑end descent state history was supplied under industry collaboration with ispace, a commercial lunar lander provider, and used as the baseline reference trajectory; the assumed mass, propulsion, and sensing specifications were cross‑checked against the current lander design capabilities to ensure engineering realism. Image sequences for the end-to-end descent were then rendered in PANGU from this state history under a nadir‑pointing camera and rangefinder configuration aligned with the Moon‑center direction throughout the profile.

\subsection{Lighting Conditions and Rendering Realism} \label{subsec:lighting-conditions-details}

To simulate realistic lunar lighting conditions, the Sun’s position was fixed to a specific timestamp: 2018 Feb 20, 00:00:00 UTC. At this moment, the Sun is approximately $0.4^\circ$ above the local horizon at the selected landing site, simulating twilight conditions typical of polar landings. Such a low solar elevation angle leads to long shadows and extreme contrast, making terrain features difficult to perceive—both for human operators and automated vision systems. As a result, most raw images appear nearly black unless enhanced through preprocessing techniques such as histogram equalization. To improve visual interpretability while maintaining a realistic context, we also simulated scenes with a slightly higher solar elevation angle of $1.35^\circ$, which helped expose terrain features while still reflecting plausible descent conditions.
A different angle was instead used for the end-to-end trajectory, where the relative Sun–Moon position was chosen so that the trajectory would follow the terminator line; this was achieved using the SPICE coordinates corresponding to the following date and time: 24/09/28 03:30.

Surface reflectance was modeled using the Hapke BRDF \cite{hapke1981bdrf}, with parameters selected from the PANGU documentation: single-scattering albedo \(w = 0.33\), opposition surge width \(h = 0.05\), opposition surge strength \(B_0 = 0.95\), and macroscopic roughness \(\theta = 0.05\). These parameters are optimized for visually realistic rendering in the PANGU environment, though they may differ from true lunar values due to scaling and approximation methods.

It is important to note that this simulation constitutes only an initial step toward photorealistic modeling of lunar descent. Several simplifying assumptions limit its physical realism. Notably, we use an idealized pinhole camera model, which does not capture sensor-level phenomena such as lens distortion, dynamic range limitations, blooming, noise, or rolling shutter effects. Furthermore, we currently neglect secondary illumination sources such as Earthshine or surface-reflected light, which may be non-negligible in low-Sun conditions, especially near the lunar poles.

To partially address these limitations, synthetic noise has been added to the rendered image sequences to better simulate degraded sensing conditions. The effect of this synthetic degradation on navigation performance is analyzed in the sensitivity analyses presented in \cref{sec:sensitivity-analysis}. Nonetheless, full photorealism would require incorporating more advanced camera models, sensor-specific calibration, real-time exposure dynamics, and multi-source lighting—all of which remain important directions for future work aimed at bridging the gap between simulation and operational deployment.

\subsection{Error Metric}\label{sec:relative-error-metric}

The relative velocity error was chosen as a robust metric for quality of egomotion estimation. This metric facilitates clearer comparison of performance in the varying regimes of the lunar landing; high altitudes and high velocity present in transfer maneuvers and braking phases of a lunar landing, in addition to the low altitudes and low velocity landing sequences. The metric is defined as:

\begin{equation}
\frac{\|\mathbf{v_{est}}-\mathbf{v_{true}}\|}{\|\mathbf{v_{true}} \| },
\label{eq:error}
\end{equation}
where $\mathbf{v_{est}}$ is the estimated velocity and $\mathbf{v_{true}}$ is the ground truth velocity.

\subsection{Computing System}

The computations in this study were performed on an Ubuntu 24.04.1 LTS virtual machine equipped with an Intel Xeon Platinum 8462Y+ 2800 MHz, 4 GB RAM. Although hardware performance was not the focus of this work, preliminary single‑threaded tests indicated throughputs on the order of tens of iterations per second at a resolution of $1024 \times 1024$. The implementation was primarily written in Python and did not undergo performance optimization. These results therefore suggest that, despite the use of a conventional general‑purpose processor and an unoptimized Python‑based pipeline, the method exhibits encouraging computational performance. Accounting for the possibility of further optimization, particularly through language‑level and algorithmic improvements, this approach holds promise for real‑time operation, though dedicated testing on the specific space‑qualified hardware targeted would be required to fully assess its feasibility.

\subsection{Algorithm Parameters}

A sensitivity analysis was completed for the optical flow to egomotion framework, outlined in \cref{sec:sensitivity-analysis}. This study involved optimization of parameters to balance accuracy and computational efficiency. For all the results depicted in this study, a resolution of $1024 \times 1024$ was used for generation of images, at a frame-rate of four Hz. Parameters for the Lucas-Kanade optical flow estimation were chosen such that points of estimation were relatively spread across the full image, averaging around 200 tracked features at any one time. Specific Lucas-Kanade and Shi-Tomasi parameters can be found in \cref{sec:lucas-kanade-params}.

\section{Results}
\label{sec:results}

In this section, we evaluate the proposed methodology, as described in \cref{sec:methodology}, on landing scenarios, Hohmann transfers, and the end-to-end trajectory introduced in \cref{sec:trajectory}. The planar model depth map, defined in \cref{eq:planar-depth}, is used for the four landing trajectories, while the sphere approximation depth map, defined in \cref{eq:spherical-depthmap}, is used for the Hohmann transfers and end-to-end trajectory. The resulting relative velocity error, as defined in \cref{eq:error}, is shown in \cref{tab:trajectory-results} which serves as high-level overview of the method performances. Here, the parameters $\alpha$, $\beta$ and $\gamma$ are computed from the camera angles rather than being estimated during the motion field inversion, which means the surface is assumed to be orthogonal to the local spheroid normal.

\begin{table}
\centering
\renewcommand\cellalign{c} 
\caption{Summary of velocity errors across all simulated trajectories. Landing trajectories use the planar depth model, while the Hohmann and end-to-end cases rely on the spherical depth approximation.}
\begin{tabular}{
    >{\centering\arraybackslash}m{2.2cm}   
    >{\centering\arraybackslash}m{2.5cm}   
    >{\centering\arraybackslash}m{3.2cm} 
    >{\centering\arraybackslash}m{1.4cm} 
    >{\centering\arraybackslash}m{1.4cm} 
    >{\centering\arraybackslash}m{1.4cm} 
    >{\centering\arraybackslash}m{1.4cm} 
}
\toprule
\multicolumn{2}{c}{\multirow{3}{*}{\makecell{Trajectory}}}
& \multirow{2}{*}{\makecell{Mean Absolute\\Velocity Error ($\pm$ m/s)}}
& \multicolumn{4}{c}{Relative Velocity Error} \\
\cmidrule(lr){4-7}
& & & Mean & Max & Min & STD \\
\midrule
\multirow{4}{*}{\makecell{Landing}} 
 & Flat          & 1.4938   & 0.0292 & 0.3890   & 0.0084 &   0.0325   \\
 & Peak          & 2.0698   & 0.0332 & 0.3533   & 0.0018 &   0.0345   \\
 & Crater        & 1.2536   & 0.0237 & 0.3203   & 0.0031 &   0.0286    \\
 & Incline       & 4.8199   & 0.0736 & 0.3631   & 0.0211 &   0.0453   \\
\midrule
Orbital & Hohmann Transfer & 28.9054 & 0.0179   & 0.0682 & 0.0001    &   0.0131   \\
\midrule
End-to-End & Full Transfer to Landing & 10.2259 & 0.0163   & 1.1492   &  0.0002   &  0.0172    \\
\bottomrule
\end{tabular}
\label{tab:trajectory-results}
\end{table}

\begin{figure}
    \centering
    \includegraphics[width=1.0\linewidth]{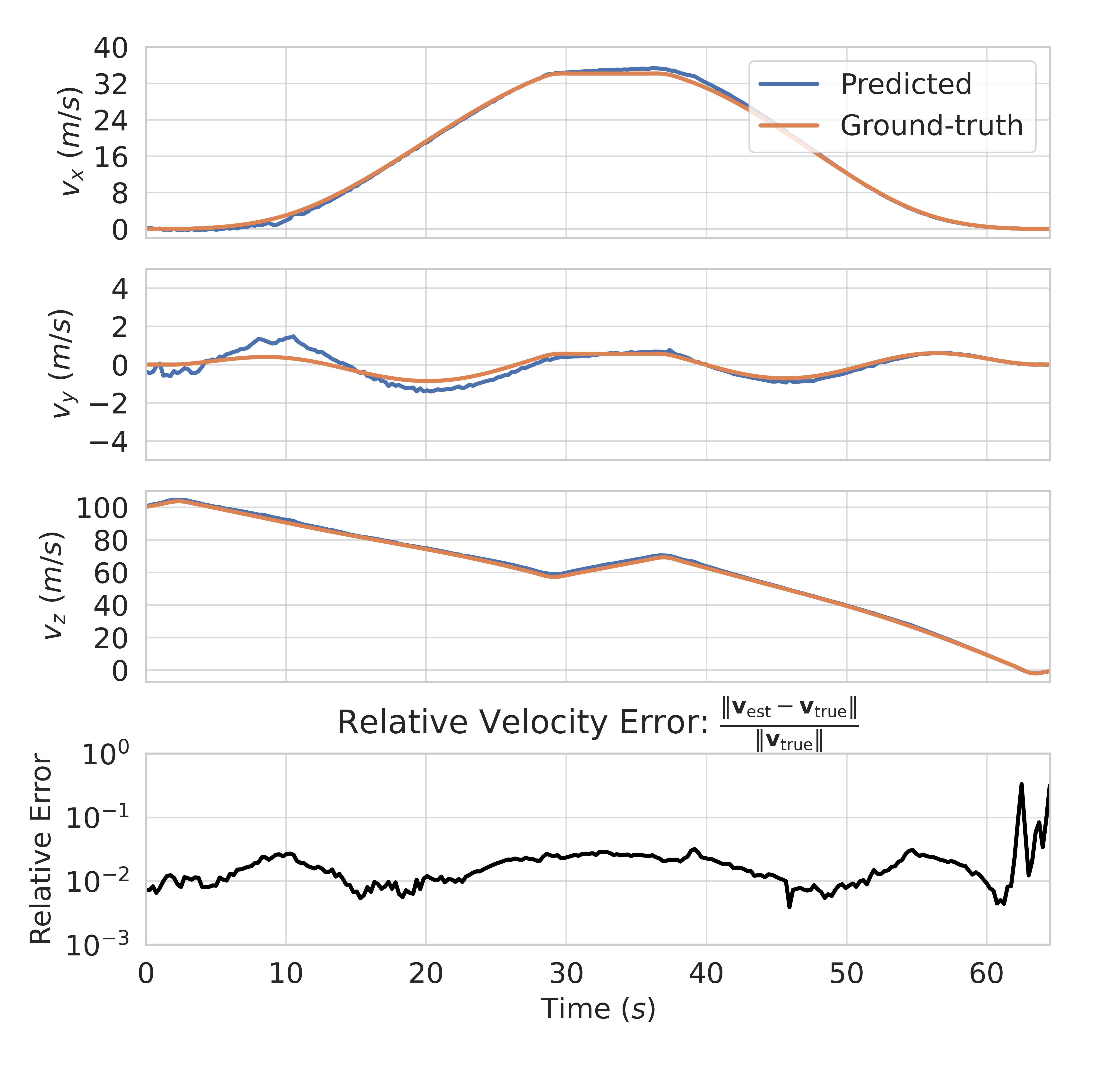}
    \caption{Comparison of estimated and ground-truth velocity components during the landing descent at the \textit{Flat} landing site, along with the relative error in velocity magnitude. Velocity estimates closely follows the ground truth throughout the descent. A pronounced error spike appears near touchdown because the ground-truth velocity magnitude approaches zero, amplifying small absolute deviations without reflecting an actual decline in performance.}\label{fig:vc_landing}
\end{figure}

A more detailed look at the results on landing scenarios is shown in \cref{fig:vc_landing} where the estimated and ground-truth velocities for the landing descent at the \textit{Flat} landing site are compared. The relative error is largely constant during the trajectory, remaining between $10^{-1}\text{ - }10^{-2}$, with the only exception being during the final seconds, where the error spikes. This final spike in the relative error is due to the ground truth velocity approaching zero, and not to degraded performance of the algorithm.

\begin{figure}
    \centering
    \subfigure[Fixed Planar Model]{
        \label{fig:ec_as_landing_site_planar}
        \includegraphics[width= 0.45\linewidth]{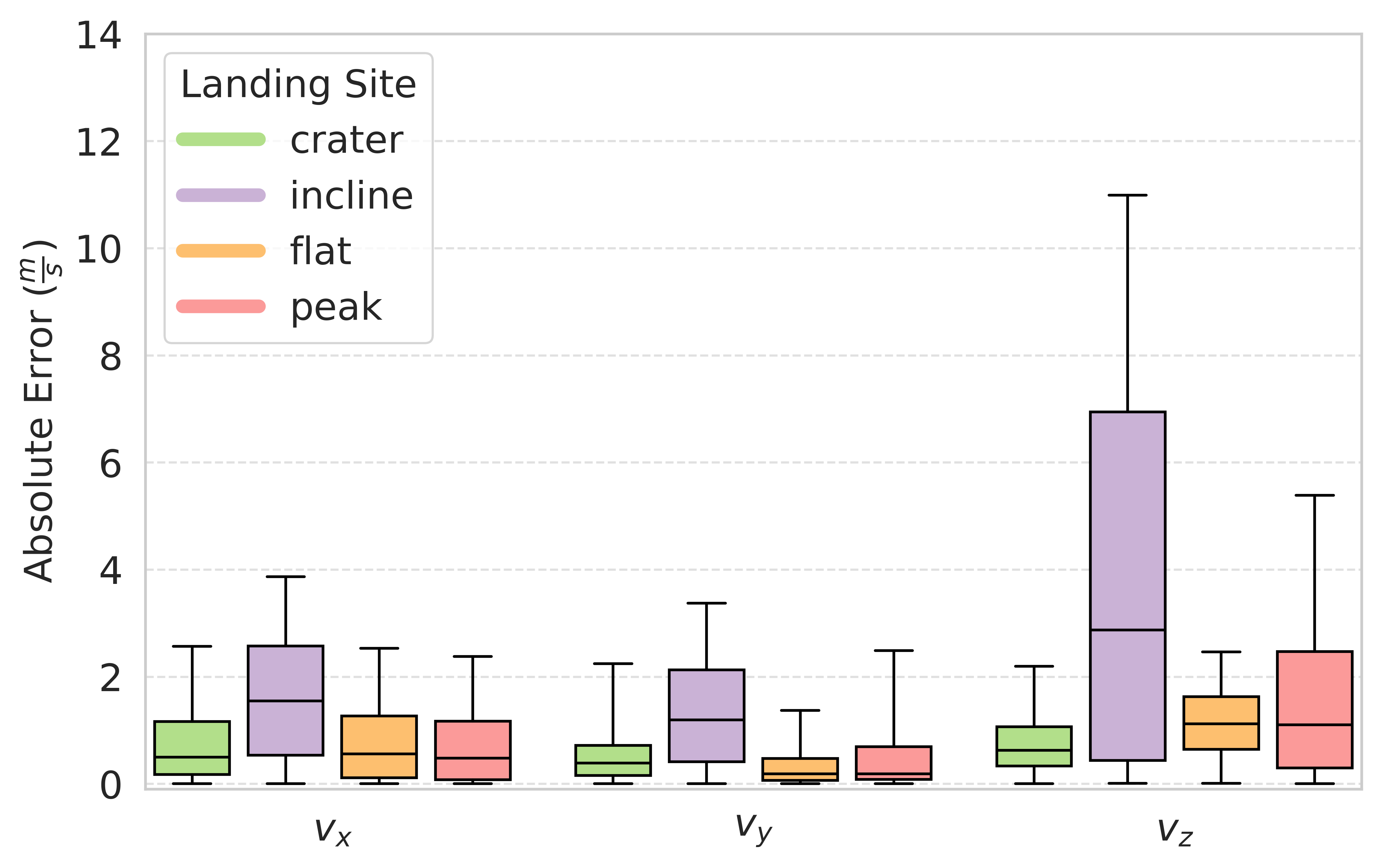}
    } 
    \subfigure[Slope Estimated Planar Model]{
        \label{fig:ec_as_landing_site_slope_est}
        \includegraphics[width= 0.45\linewidth]{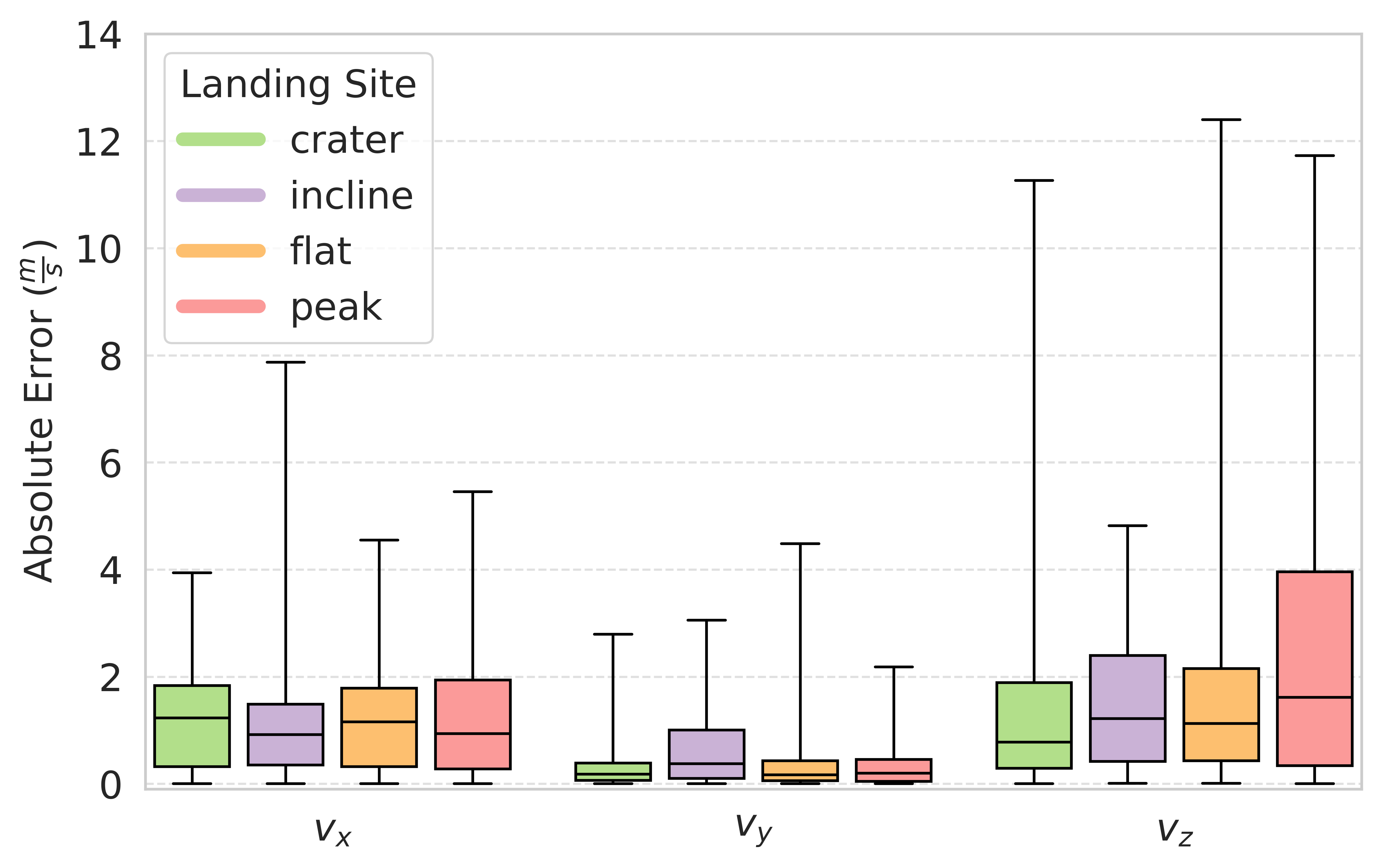}
    } 
    \caption{Boxplots of velocity estimation error for different landing sites using the \textit{fixed} planar model on the one hand in (a), and the \textit{slope estimated} planar model on the other hand in (b). }
    \label{fig:ec_as_landing_site_comparison}
\end{figure}

\cref{tab:trajectory-results} shows how the \textit{Crater} landing site velocity estimation achieves a similar error to the \textit{Flat} landing site, both being located in a similarly flat region of the Moon. With some degradation, the \textit{Peak} landing site velocity estimation also shows an error in the same order of magnitude. The velocity estimation over the \textit{Incline} landing site presents a higher error, as it is placed in regions with slopes and inclined terrain, conflicting with the orthogonal surface assumption. This conflict can be remediated using the \textit{slope estimation} planar model in which $\alpha$ and $\beta$ are estimated alongside the translational velocity. As described in \cref{subsec:motion-field-inversion}, the formulation becomes a non-linear least squares problem, which increases computational complexity and can introduce optimization instabilities. Results of this experiment, as shown in \cref{fig:ec_as_landing_site_comparison}, indicate an improved $v_z$ estimation above the \textit{Incline} site, where the fixed planar model deteriorated, at the cost of a slightly degraded performance and greater variance of the velocity estimation error across the more uniform landing sites. These findings suggest that while slope estimation enhances robustness to non-uniform terrain, it reduces overall reliability of the optimization algorithm used in this work. Different optimization methods could improve the sensitivity of the velocity estimation error to the increase in optimization complexity.

For the trajectories featuring higher altitudes, namely the Hohmann transfers and the end-to-end trajectory, the absolute error is higher than in the the landing trajectories due to higher orbital velocities involved, however, the relative error is lower. The lowered relative error is probably due to local terrain variations being negligible relative to the spacecraft altitude, making the spherical depth model approximation a better approximation at higher altitudes than the fixed planar depth model is at lower altitudes. For both of these trajectories, the mean error in \cref{tab:trajectory-results} is below $0.02$, showing the applicability of the proposed approach in a broad range of altitudes up to 300 km.

\begin{figure}
    \centering
    \includegraphics[width=1.0\linewidth]{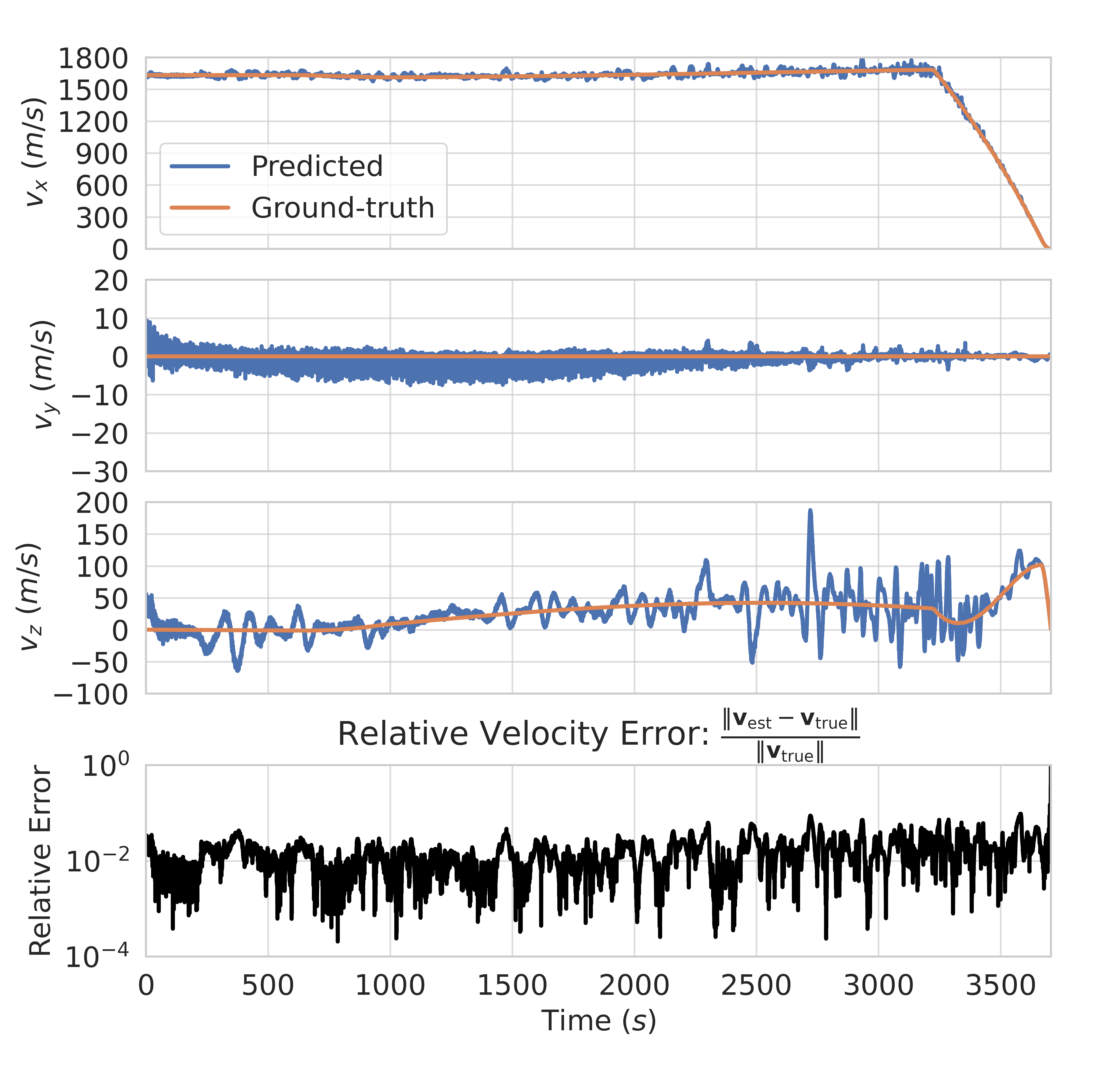}
    \caption{Estimated versus ground-truth velocity components over the full end-to-end lunar landing trajectory provided by ispace, with corresponding relative velocity error. The method maintains accuracy across all phases, from high-altitude orbital approach to terminal descent, with the relative error remaining in the order of $10^{-2}$ despite changes in altitude, velocity, and illumination. A sharp spike in the error occurs near touchdown due to the ground-truth velocity magnitude approaching zero, which amplifies minimal absolute deviations without indicating performance degradation.}
    \label{fig:ispace-trajectory-error-comparison}
\end{figure}

\Cref{fig:ispace-trajectory-error-comparison} shows a comparison between the ground-truth and estimated velocity through the whole end-to-end trajectory, as well as the resulting error, which remains around the order of $10^{-2}$ across the full trajectory, despite large variations in altitude, velocity, and lighting conditions. The translational velocity estimates remained consistently accurate, while the vertical velocity showed larger fluctuations. These fluctuations likely stem from rangefinder sensitivity to rapid translational motion over varying terrain, suggesting that more frequent polling or improved filtering could mitigate such effects. Overall, the results demonstrate that the proposed method can deliver reliable velocity estimation across all major phases of a lunar landing trajectory.

\section{Sensitivity and Robustness Analyses}\label{sec:sensitivity-analysis}

\begin{figure}
    \centering
    \subfigure[Effect of Spatial Resolution]{
        \includegraphics[width=0.45\linewidth]{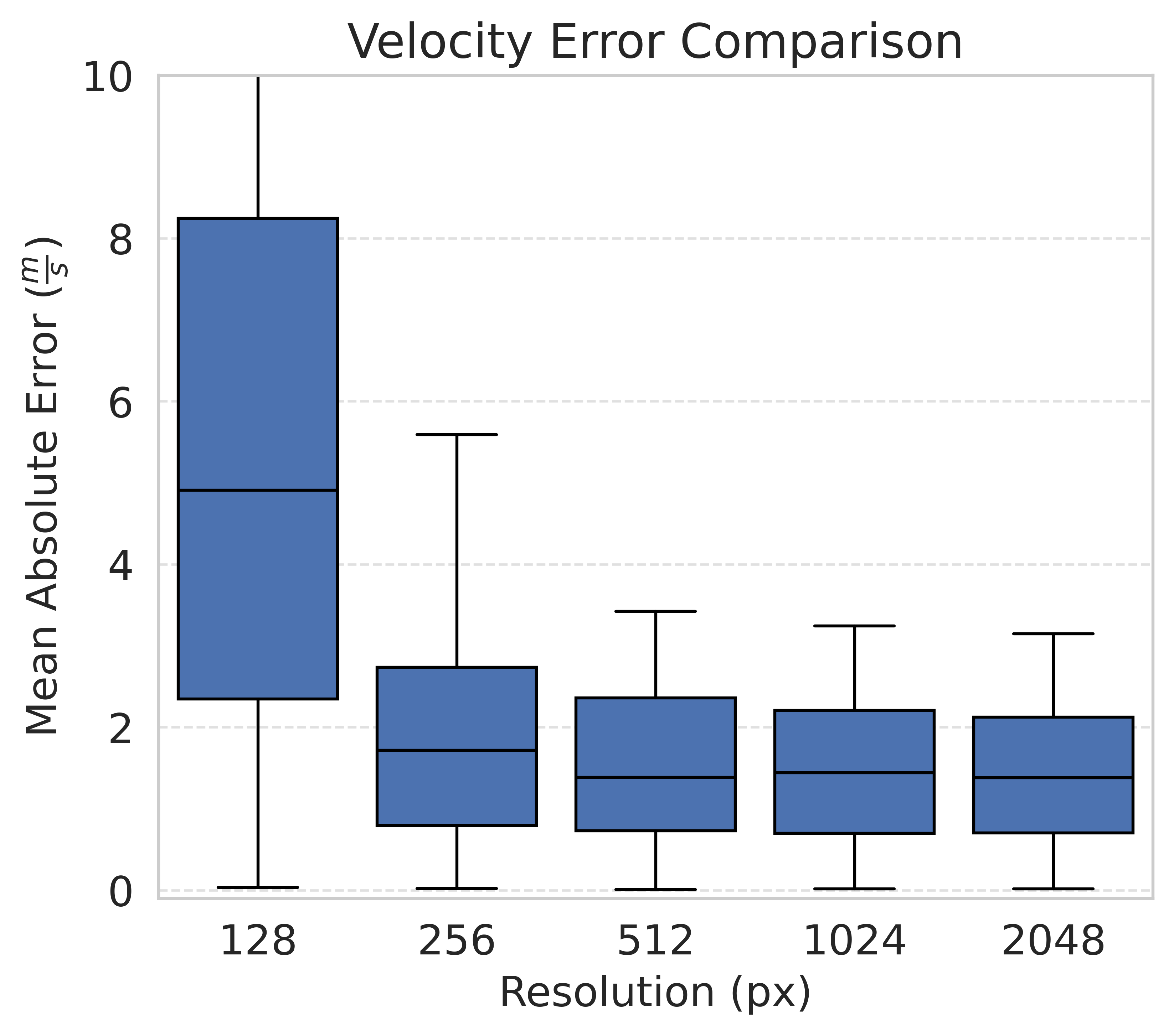}
        \label{fig:ec_as_spatial}
    }
    \subfigure[Effect of Temporal Resolution]{
        \includegraphics[width=0.45\linewidth]{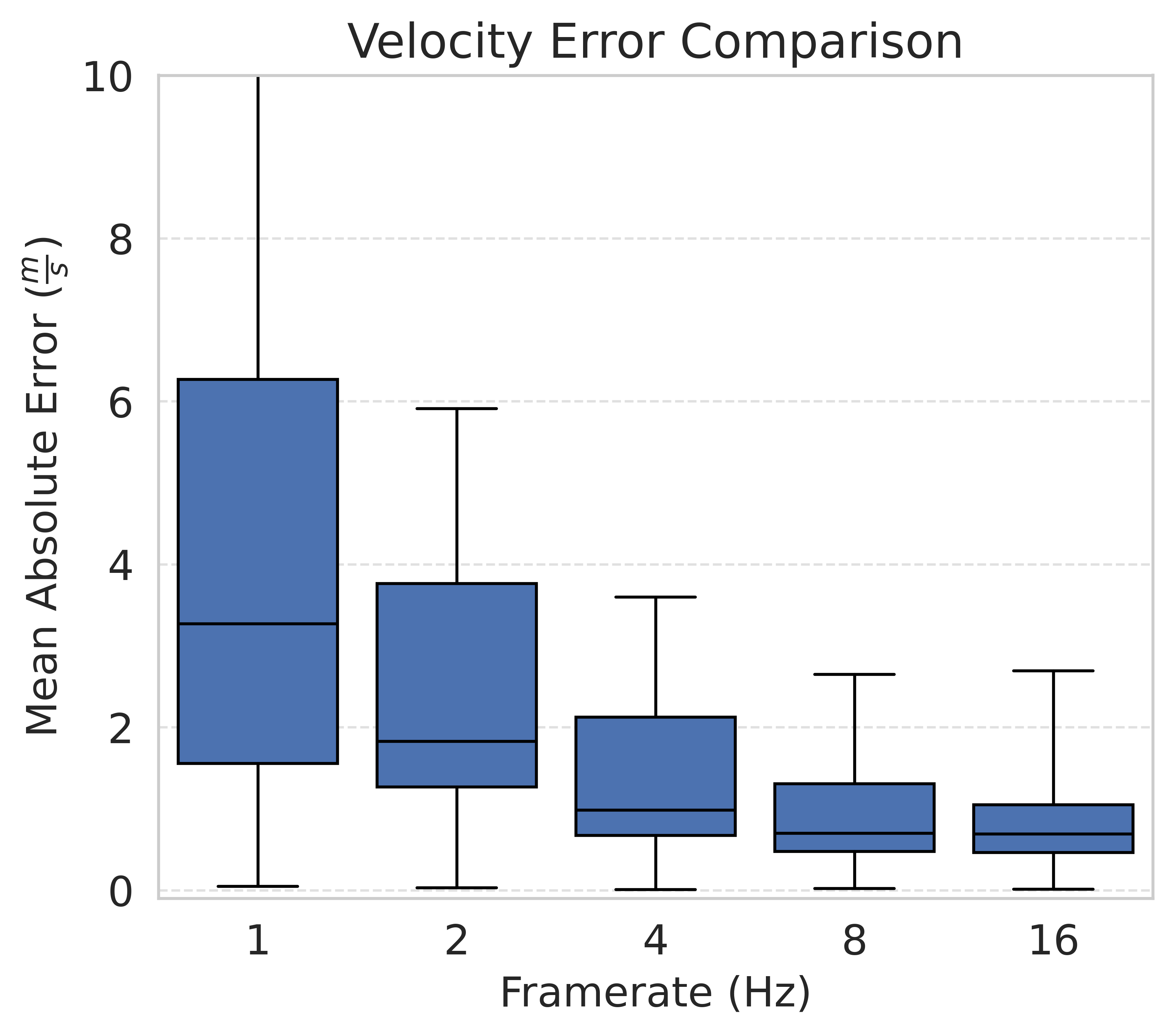}
        \label{fig:ec_as_temporal}
    }\\
    \subfigure[Effect of Camera Noise]{
        \includegraphics[width=0.45\linewidth]{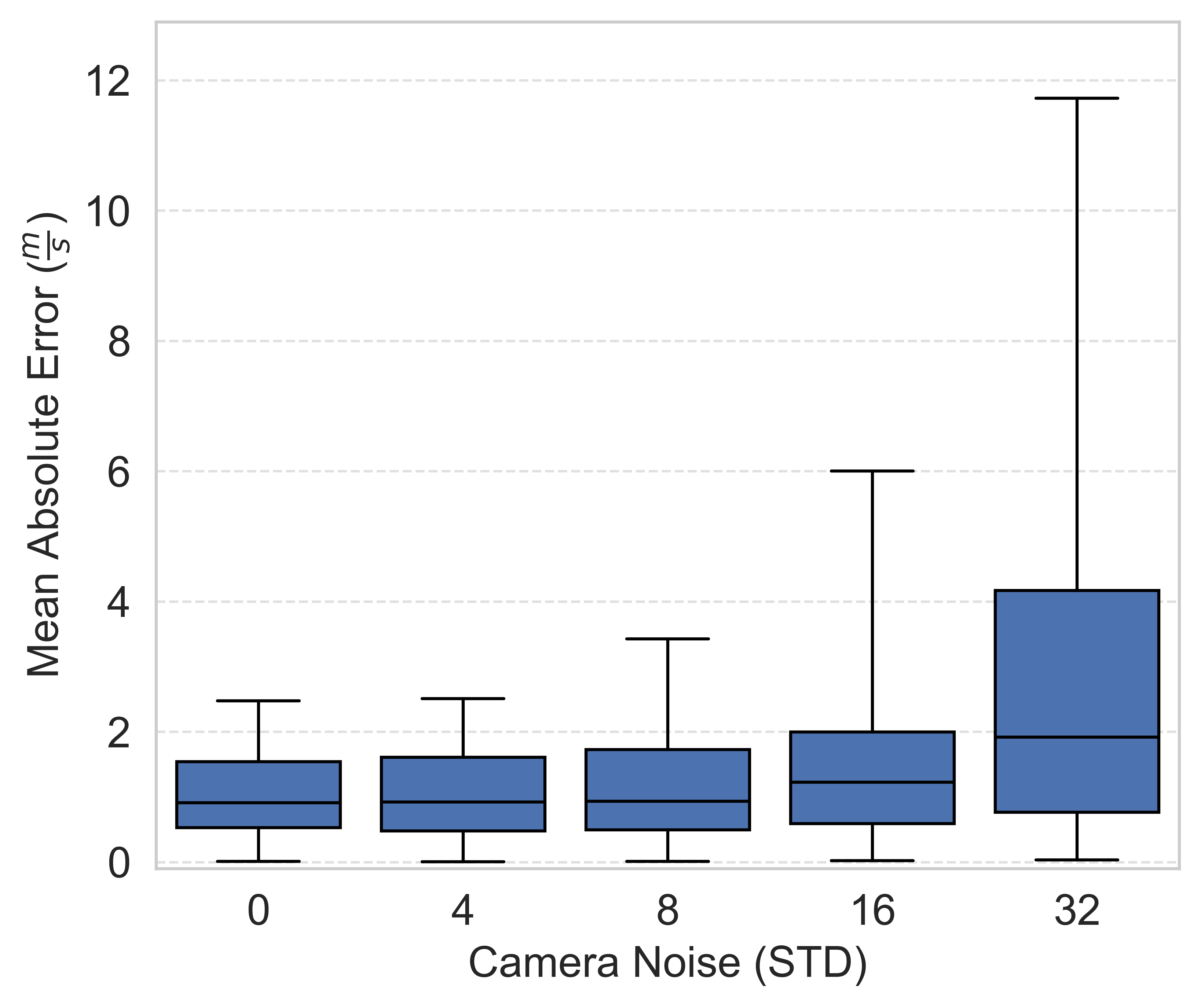}
        \label{fig:ec_as_camera_noise}
    }
    \subfigure[Effect of State Noise]{
        \includegraphics[width=0.45\linewidth]{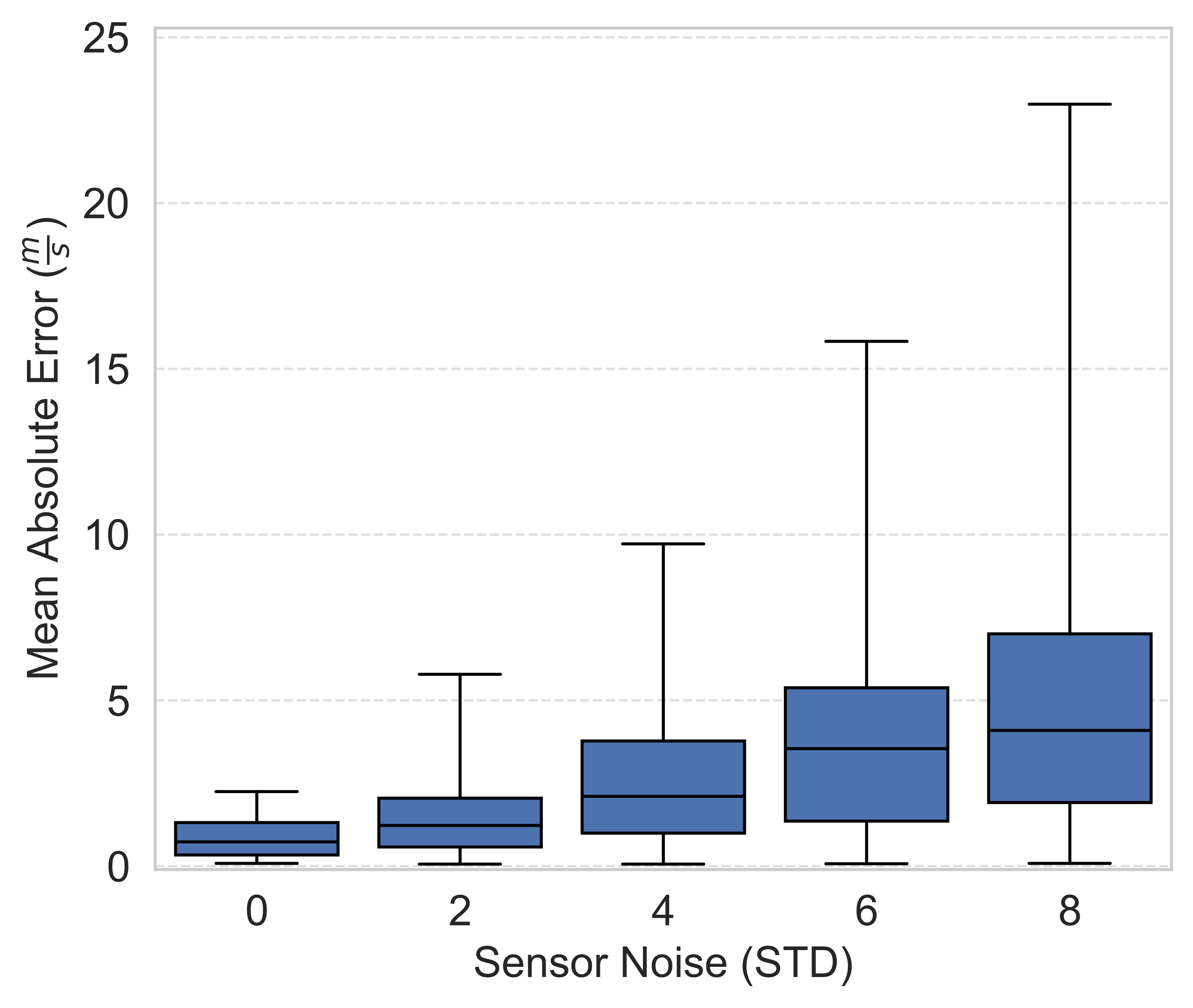}
        \label{fig:ec_as_state_noise}
    }
    \caption{Velocity error comparison at the \textit{Flat} landing site. (a) and (b) show the influence of spatial and temporal resolution on egomotion estimation accuracy, respectively. (c) and (d) show the influence of artificial noise in the camera and state, respectively, on the egomotion estimation accuracy.}
    \label{fig:ec_as_2x2}
\end{figure}

To assess the robustness of the proposed framework, a series of analyses were performed, evaluating how different factors affect performance: spatial resolution, temporal resolution, camera noise, and state noise. Understanding these sensitivities can help define more optimal system parameters for future implementations. \Cref{fig:ec_as_2x2} shows how varying the aforementioned settings affects the landing trajectory performance, which, as discussed in \cref{sec:results}, represents the most challenging scenario for the methodology proposed in this study. The \textit{Flat} landing site was selected to minimize depth map-related errors. The rest of this section analyses in detail the effect of the individual parameters on performance.

\subsection{Spatial Resolution}\label{subsec:sa-spatial-resolution}

As shown in \cref{fig:ec_as_spatial}, increasing the resolution causes the error to drop until the gain saturates at $1024 \times 1024$ pixels. This decrease in the erro is expected, as increasing the resolution allows for a more accurate tracking of features by the Lucas-Kanade algorithm, thereby reducing the optical flow error.
In case of hardware or computational limitations, lower resolutions of 512 or even 256 pixels per side remain viable with only a marginal error increase, but these lower resolutions may be more susceptible to noisy, low-light scenarios.

An important consideration is that the feature tracking by the optical flow estimation is fundamentally restricted by the detail in the images, which in turn is capped by the resolution of the textures themselves in the synthetic PANGU model. In other words, even if the resolution were to increase substantially, the feature tracking would still reach saturation due to the lack of detail in the textures of the synthetic PANGU model. To remedy this, efforts were made to ensure realistic models, which are outlined in \cref{subsec:lighting-conditions-details}.

\subsection{Temporal Resolution}\label{subsec:sa-temporal-resolution}

A lower temporal resolution (i.e. a reduced camera frame-rate) increases the time interval over which optical flow and velocity estimations are interpolated. As illustrated in \cref{fig:ec_as_temporal}, this produces a strong inverse relationship between frame-rate and estimation error, with a performance impact even greater than that of spatial resolution.

The underlying cause of this dependence lies in the temporal continuity of the motion field equations, in contrast to the discrete nature of optical flow estimation. Since optical flow is computed between image frames separated by a finite time step, the result corresponds to the average motion over that interval rather than the instantaneous motion field. Similarly, IMU state measurements, sampled asynchronously with respect to the camera, are interpolated between frames, introducing analogous averaging artifacts. These effects become more pronounced in scenarios with rapid variations in angular or linear velocity.

Beyond these theoretical aspects, temporal resolution also affects feature tracking performance. At lower frame-rates, the displacement of visual features between consecutive frames increases, making it more difficult for the Lucas–Kanade algorithm to match features consistently, particularly under the highly dynamic conditions encountered during low-altitude descent. Increasing the frame-rate reduces inter-frame motion, thereby improving the stability and robustness of optical flow computation.

The combined impact of temporal averaging effects and feature-tracking limitations explains the pronounced sensitivity to temporal resolution observed in the results. Empirically, a frame-rate of four Hz provided a suitable trade-off between accuracy and computational cost for the lunar descent scenario considered. Nevertheless, higher frame-rates would likely enhance performance further, especially during high-dynamics phases such as hazard-avoidance maneuvers. In contrast, for trajectories with a ventral camera orientation and limited angular velocity changes, such as the one analyzed in \cref{subsec:temporal-resolution-study-ispace}, the dependence on frame-rate was noticeably weaker, as the reduced nonlinear motion within frame intervals diminished both averaging effects and tracking difficulties.

\subsection{Camera Noise}\label{subsec:sa-camera-noise}

Camera noise was modeled by independently sampling each pixel intensity from a Gaussian distribution $\mathcal{N}(0,\sigma)$, with tested standard deviations (STD) up to 32, corresponding to one-eighth of the full pixel intensity range. As illustrated in \cref{fig:ec_as_camera_noise}, the proposed pipeline demonstrated strong resilience to substantial levels of camera noise. Performance degradation became significant only for extreme noise levels above 16~STD. This behavior is expected, as uncorrelated pixel noise tends to average out across the large number of tracked features, particularly at higher spatial resolutions, thus limiting its impact on both optical flow computation and egomotion estimation accuracy.

\subsection{State Noise}\label{subsec:sa-sensor-noise}

In this study, Gaussian noise was added to the IMU states. It can be seen in \cref{fig:ec_as_state_noise} that the framework is less robust to state noise than to camera noise. This reduced resilience is expected, as these state variables are singular measurements whose errors cannot be mitigated by spatial averaging without the use of filtering. While the random errors caused by state noise clearly impact egomotion estimation, systematic errors and in particular long-term sensor drift represent an even more significant risk for autonomous landers relying on propagated IMU states, though this aspect lies beyond the scope of the present work.

In addition, the larger sensitivity is caused by the fact that small angular velocities can generate large apparent motion in the image. The effect of this phenomenon must be minimized to isolate the optical flow induced by translational movement. This phenomenon becomes particularly pronounced when the translational component of the flow is small. Accurate measurements of attitude and angular velocity from the IMU are therefore even more critical.




\section{Conclusions}

This study demonstrates that vision-based egomotion estimation, combining sparse optical flow with simplified depth modeling, can provide a reliable and computationally efficient solution for autonomous lunar landing navigation. Both planar and spherical depth approximations, when complemented by rangefinder measurements, were found to yield accurate velocity estimates across varying altitudes and terrain profiles. The proposed approach operates within modest computational budgets, making it suitable for deployment on the resource-limited platforms characteristic of small lunar landers.

The sensitivity analyses revealed a pronounced dependence on temporal resolution and state accuracy, highlighting the importance of balancing hardware limitations with real-time navigation accuracy. A typical onboard camera configuration (resolution $\geq$512~px, frame-rate $\geq$4~Hz), combined with a rangefinder and an accurate IMU, as presumed in this study, is shown to be sufficient to support reliable optical-flow–based egomotion estimation. Compared to commonly used LiDAR-based alternatives that are typically heavy and power-demanding \cite{amzajerdian2018navigation, pierrottet2009flight}, the proposed approach offers a far more lightweight and energy-efficient solution, relying solely on onboard computation and potentially pre-existing cameras.

Preliminary tests with a RANSAC-based variant of the least-squares estimation did not yield notable improvements; nevertheless, introducing appropriate filtering mechanisms could mitigate optical flow estimation errors expected in real flight conditions. Furthermore, integrating this framework with sensor fusion methods such as Kalman filtering \cite{verweld2013relative} may improve the accuracy of attitude and angular velocity estimates, enhancing stability and robustness in operational scenarios.

Overall, the results establish the practicality and effectiveness of this lightweight vision-based framework, advocating for its integration into future governmental and commercial lunar missions where autonomy, reliability, and efficiency remain central design objectives.

\section*{Author contributions}

SC, PF and LW contributed equally to this work and to writing the paper. DI supervised all the conducted experiments and oversaw the preparation of the manuscript. CHY, KA and YN provided some insights on realistic needs and constraints of a private lunar landing mission.
All authors read and approved the final manuscript.

\section*{References}

\bibliographystyle{astrobib}
\bibliography{refs}

\appendix
\section{Appendix}

\subsection{Temporal Resolution study for end-to-end trajectory}\label{subsec:temporal-resolution-study-ispace}

The influence of temporal resolution on egomotion accuracy for the end-to-end trajectory is summarized in Table~\ref{tab:ispace-temporal-resolution-results}. The results show stable performance down to 1~Hz, with a sharp degradation observed for lower frame-rates. The indifference to frame-rates above 1~Hz highlights that the averaging effects, described in Section~\ref{subsec:sa-temporal-resolution}, are less prevalent when motion is more linear, with degraded performance becoming challenging only below 1~Hz -- where feature tracking begins to struggle.

\begin{table}[H]
\centering
\renewcommand\cellalign{c} 
\caption{Egomotion estimation error statistics for different temporal resolutions (frame-rates).}
\begin{tabular}{
    >{\centering\arraybackslash}m{2.5cm}   
    >{\centering\arraybackslash}m{3.2cm}   
    >{\centering\arraybackslash}m{1.4cm}   
    >{\centering\arraybackslash}m{1.4cm}   
    >{\centering\arraybackslash}m{1.4cm}   
    >{\centering\arraybackslash}m{1.4cm}   
}
\toprule
\multirow{2}{*}{\makecell{Frame-rate \\ (Hz)}}
& \multirow{2}{*}{\makecell{Mean Absolute \\ Pose Error ($\pm$ m)}}
& \multicolumn{4}{c}{Relative Pose Error} \\
\cmidrule(lr){3-6}
& & Mean & Max & Min & STD \\
\midrule
4.0   & 10.2259 & 0.0163 & 0.0172 & 1.1492 & 0.0002 \\
2.0   & 9.8499 & 0.0159 & 0.0150 & 0.4625 & 0.0002 \\
1.0   & 9.7195 & 0.0165 & 0.0341 & 1.7517 & 0.0002 \\
0.5   & 12.8536 & 0.0234 & 0.1264 & 5.2228 & 0.0002 \\
0.25  & 89.9750 & 0.1454 & 0.2616 & 1.0849 & 0.0003 \\
0.125 & 277.9690 & 0.4620 & 0.4162 & 2.4615 & 0.0007 \\
\bottomrule
\end{tabular}
\label{tab:ispace-temporal-resolution-results}
\end{table}

\subsection{Lucas-Kanade with Shi--Tomasi feature selection and pyramidal feature tracking}\label{sec:lucas-kanade-params}
Table~\ref{tab:lk-shitomasi-params} lists the parameters employed for feature detection and tracking, combining Shi--Tomasi corner extraction with Lucas--Kanade optical flow using a pyramidal scheme to ensure robust tracking across scales.

\begin{table}[H]
\centering
\renewcommand\cellalign{c} 
\caption{Parameters used for Shi--Tomasi feature selection and Lucas--Kanade pyramidal optical flow tracking.}
\begin{tabular}{
    >{\centering\arraybackslash}m{4.5cm}  
    >{\centering\arraybackslash}m{4.5cm}  
    >{\centering\arraybackslash}m{3.0cm}  
}
\toprule
\multicolumn{2}{c}{\multirow{2}{*}{\makecell{Feature Detection \\ \& Tracking Parameters}}}
& \multirow{2}{*}{\makecell{Value}} \\
\\
\midrule
\multirow{4}{*}{\makecell{Shi--Tomasi \\ Corner Detection}}
    & Max. Corners ($N_\text{max}$)  & 1000 \\
    & Quality Level ($q$)            & 0.1 \\
    & Min. Distance ($d_\text{min}$) & 50 px \\
    & Block Size ($b$)               & 10 px \\
\midrule
\multirow{3}{*}{\makecell{Lucas--Kanade \\ Optical Flow (with pyramids)}}
    & Window Size ($w \times h$)     & $50 \times 50$ px \\
    & Max. Pyramid Levels ($L$)      & 4 \\
    & Termination Criteria ($\epsilon$, $N_\text{iter}$) & $\epsilon = 0.03$, $N=10$ \\
\bottomrule
\end{tabular}
\label{tab:lk-shitomasi-params}
\end{table}


\subsection{Spherical Depth Map Further Derivation}

An extra step that is non-trivial is relating $\alpha$, $\beta$, and $\gamma$ to the attitude of the spacecraft. This allows for the analytical determination of $\alpha$, $\beta$, and $\gamma$, which subsequently don't have to be learned by the least squares estimation discussed in \cref{subsec:motion-field-inversion}.
Given $\phi$, $\theta$, and $\psi$ as the angles rotating the spacecraft from a body frame to the camera frame (given in \cref{fig:sphere}) around the X, Y, and Z axis, respectively. Then a rotation matrix $\textbf{R}^C_B$ can be formed . The body frame (with superscript $b$) is defined to be one where the positive z-axis points towards the center of the moon, the cross product between the z-axis and the velocity vector gives the y-axis, and finally the cross-product between the y- and z-axis give the x-axis. The angle $\nu$ can be found by$$\mathbf{\hat{z}^C} = \mathbf{R}^C_B \mathbf{\hat{z}^B},$$ giving an expression for the z-axis in the camera. Now $$\mathbf{\hat{z}^C} \cdot \mathbf{\hat{z}^B} = \cos \nu$$ gives $\nu$. Subsequently applying the sine rule for the triangle MPO, one can find $\mu$ with $$\frac{\sin \nu}{R} = \frac{\sin \mu}{\rho}.$$ Using the sum of angles for a triangle, one gets $$\lambda = \pi - \mu - \nu,$$ and finally H can be found with a second application of the sine rule giving $\frac{\sin \lambda}{H + R} = \frac{\sin \nu}{R}$.

Using previously constructed vectors, $\hat{k}^c$ can now be found:

\begin{align}
    \hat{k}^c &= \frac{\vec{r}^{\, c}_{OP} - \vec{r}^{\, c}_{OM}}{||\vec{r}^{\, c}_{OP} - \vec{r}^{\, c}_{OM}||} \\
    &= \frac{\begin{bmatrix}
        0, 0, \rho
    \end{bmatrix}^T - (H + R) \mathbf{R}^C_B \mathbf{\hat{z}^B}}{||\begin{bmatrix}
        0, 0, \rho
    \end{bmatrix}^T - (H + R) \mathbf{R}^C_B \mathbf{\hat{z}^B}||}
\end{align}

\subsection*{Author biography}

\begin{biography}[sean]{Sean Cowan} obtained a Master's degree in Aerospace Engineering following a Bachelor's degree in Mechanical Engineering at Delft University of Technology. During his time there, he worked as teaching assistant for the course 'Propagation and Optimization in Astrodynamics' and as technical assistant on the development of an in-house astrodynamics toolbox; Tudat. After graduating, Sean worked in Turin at Ecosmic, where he was the core developer of their first main product aimed at improving the Space Traffic Management capabilities of satellite operators. He then joined the European Space Agency's Advanced Concepts Team as a Young Graduate Trainee in Advanced Mission Analysis. Here, he is conducting research on various spaceflight dynamics problems, among others on verified integration techniques for GNC applications.
\end{biography}

\begin{biography}[pietro]{Pietro Fanti} holds a Master’s degree in Artificial Intelligence from the University of Bologna (Italy) and a Bachelor's degree in Computer Science from the University of Florence. He currently works at the European Space Agency’s Advanced Concepts Team. His research integrates AI techniques with space sciences to address complex problems in gravitational modeling and spacecraft dynamics. Prior to this, Pietro gained experience in applied AI research in industry, focusing on generative computer vision and developing scalable AI solutions for widely used applications.
\end{biography}

\begin{biography}[leon]{Leon B. S. Williams} completed an integrated Master's degree in Chemical Physics with work placement at the University of Glasgow. During his time at university, Leon spent 1 year working at Diamond Light Source synchrotron, in Oxford, where he carried out advanced materials science research, studying novel synthesis methods for Graphene and other 2D materials in Ultra-High Vacuum using high-intensity synchrotron radiation. During his masters project, he studied agent-based modeling using kinetic theory of gases. After graduating, he studied machine and deep learning techniques before working as a ML Engineer at Sonomatic Ltd, developing ML methods for ultrasonic testing. Since then, Leon has joined the European Space Agency, where he works in the Advanced Concepts Team carrying out open research on future space technologies, and hosting open competitions. 
\end{biography}

\begin{biography}[hippo]{Chit Hong Yam} received his Ph.D. from Purdue University. He subsequently served as a postdoctoral researcher at the European Space Agency (ESA) and the Japan Aerospace Exploration Agency (JAXA), where he supported the design of several deep-space missions. He is currently affiliated with ispace, inc., where his work focuses on advancing guidance, navigation, and control methodologies for upcoming lunar exploration missions.
\end{biography}

\begin{biography}[asakuma]{Kaneyasu Asakuma} graduated as a Master's degree in theoretical physics from the Tokyo Institute of Science (Japan) (formerly Tokyo Institute of Technology). He worked at Mitsubishi Electric Software (formerly Mitsubishi Space Software), where he was engaged in the development of navigation and guidance systems and simulation software for the H-IIA, H-IIB, H3, and Epsilon launch vehicles. He currently works as the Guidance, Navigation, and Control Group Manager, overseeing orbital and descent GNC, image-based navigation systems, and the development of GNC sensors.
\end{biography}

\begin{biography}[nada]{Yuichiro Nada} holds a Master’s degree in Aeronautics and Astronautics from the University of Tokyo, Japan. During his time there, he contributed to the development of a propulsion control system for a commercial small satellite and to research on orbit determination for micro spacecraft and solar power sail technologies at the Japan Aerospace Exploration Agency (JAXA). He later worked on system-level architecture, the design of intended functions, and functional safety for autonomous-driving systems in the automotive industry. He is currently a Lead Engineer for Vision-Based Navigation (VBN) and Hazard Detection and Avoidance (HDA) at ispace inc., where he develops image-based navigation and autonomous landing functions for lunar landing missions.
\end{biography}

\begin{biography}[dario]{Dario Izzo} graduated as a Doctor of Aeronautical Engineering from the University Sapienza of Rome (Italy). He then took a second master in Satellite Platforms at the University of Cranfield in the United Kingdom and completed his Ph.D. in Mathematical Modelling at the University Sapienza of Rome where he lectured classical mechanics and space flight mechanics. Dario Izzo later joined the European Space Agency and became the scientific coordinator of its Advanced Concepts Team. He devised and managed the Global Trajectory Optimization Competitions events and the Kelvins innovation and competition platform and made key contributions to the understanding of flight mechanics and spacecraft control and pioneering techniques based on evolutionary and machine-learning approaches. Dario Izzo received the Humies Gold Medal, the Barry Carlton award and led the team winning the 8th edition of the Global Trajectory Optimization Competition.
\end{biography}




\end{document}